\documentclass[letterpaper,10pt,journal,twoside]{IEEEtran}
\usepackage{amsmath}
\usepackage{graphicx}
\usepackage{booktabs}
\usepackage{multirow}
\usepackage{tabularx}
\usepackage{array}
\usepackage[nocompress]{cite}

\usepackage[colorlinks=true,linkcolor=blue,citecolor=blue,urlcolor=blue,filecolor=blue]{hyperref}
\title{MagCilia: A Compact Magnetociliary Tactile Sensor with 3D Force Sensing for Robotic Contact Perception and Grasping Feedback}
\author{Yu~Feng,
        Hao~Wu,
        Haotian~Guo,
        Haoming~Liu,
        William~Su,
        Jingxiang~Guo,
        Jiankun~Li,
        Masayoshi~Tomizuka,~\IEEEmembership{Life Fellow,~IEEE,}
        Wen~Jung~Li,~\IEEEmembership{Fellow,~IEEE,}
        and~Jianshu~Zhou,~\IEEEmembership{Member,~IEEE}
\thanks{
This work was supported in part by the National University of Singapore under the NUS Start-up Grant (FY2026). Yu~Feng, Hao~Wu, and Haotian~Guo contributed equally. \textit{(Corresponding authors: Wen Jung Li and Jianshu Zhou.)}}
\thanks{Yu Feng, Hao Wu, Haotian Guo, Haoming Liu, Jingxiang Guo, and Jianshu Zhou are with the Department of Mechanical Engineering, National University
of Singapore, Singapore (e-mail: jianshuzhou@nus.edu.sg).}
\thanks{William Su and Masayoshi Tomizuka are with the Department of Mechanical Engineering, University of California, Berkeley, Berkeley, CA, USA.}
\thanks{Jiankun Li and Wen Jung Li are with the Mechanical Engineering Department, City University of Hong Kong, Kowloon, Hong Kong SAR, and CAS-CityU Joint Laboratory for Robotic Research, City University of Hong Kong Shenzhen Research Institute, Shenzhen, China (e-mail: wenjli@cityu.edu.hk).}
}

\begin{document}
\bstctlcite{reference-format}
\markboth{IEEE Transactions on Industrial Electronics}%
{Feng \MakeLowercase{\textit{et al.}}: MagCilia: A Magnetociliary Tactile Sensor for 3D Force Sensing}
\maketitle

\begin{abstract}
Robotic grasping and surface exploration benefit from simultaneous measurement of normal and tangential forces and from surface information obtained through contact. Here, we present a compact magnetociliary tactile sensor (MagCilia) that combines a flexible magnetic-cilia structure with a Hall sensor for 3D force sensing. Quasi-static finite element analysis is used to investigate structural deformation and magnetic responses under multidirectional loading. To reconstruct forces from the coupled magnetic channels, we propose causal history fusion regression (CHFR), which combines current magnetic-field measurements with their recent changes. Five-fold cross-validation grouped by calibration record yields root-mean-square errors of 0.40, 0.57, and 0.69 N for $F_x$, $F_y$, and $F_z$, respectively, with corresponding coefficients of determination of 0.93, 0.90, and 0.92. Robotic experiments demonstrate tangential-force-guided gripper adjustment and multi-axis load monitoring under external perturbations. Frequency-domain features of the reconstructed forces distinguish six surface categories with 99.39\% accuracy in three-fold cross-validation grouped by acquisition session. An online robotic demonstration additionally identifies all six tested surfaces. These results demonstrate 3D force reconstruction, grasping feedback, and surface recognition using a single compact tactile unit.
\end{abstract}


\section{Introduction}
\IEEEPARstart{A}{s} robots increasingly move from structured environments to complex physical interaction scenarios, grasping and manipulation place greater demands on tactile sensing \cite{Li2020review}. Contact forces characterize the mechanical interaction between the end effector and the object \cite{Park2025tactile}. The normal force indicates the degree of gripping and compression, whereas object weight, external perturbations, and relative motion at the contact interface can generate tangential loads in different directions. Simultaneous measurement of normal and tangential loads therefore provides a more complete description of the contact interaction and supports stable grasping and precise manipulation \cite{Mo2026highly}. Tactile and multimodal sensing have also been used to adjust grasping based on the sensed contact state \cite{Su2024soft,Wang2026relaxation}.
 
\begin{figure}[t]
\centering
\includegraphics[width=0.85\columnwidth]{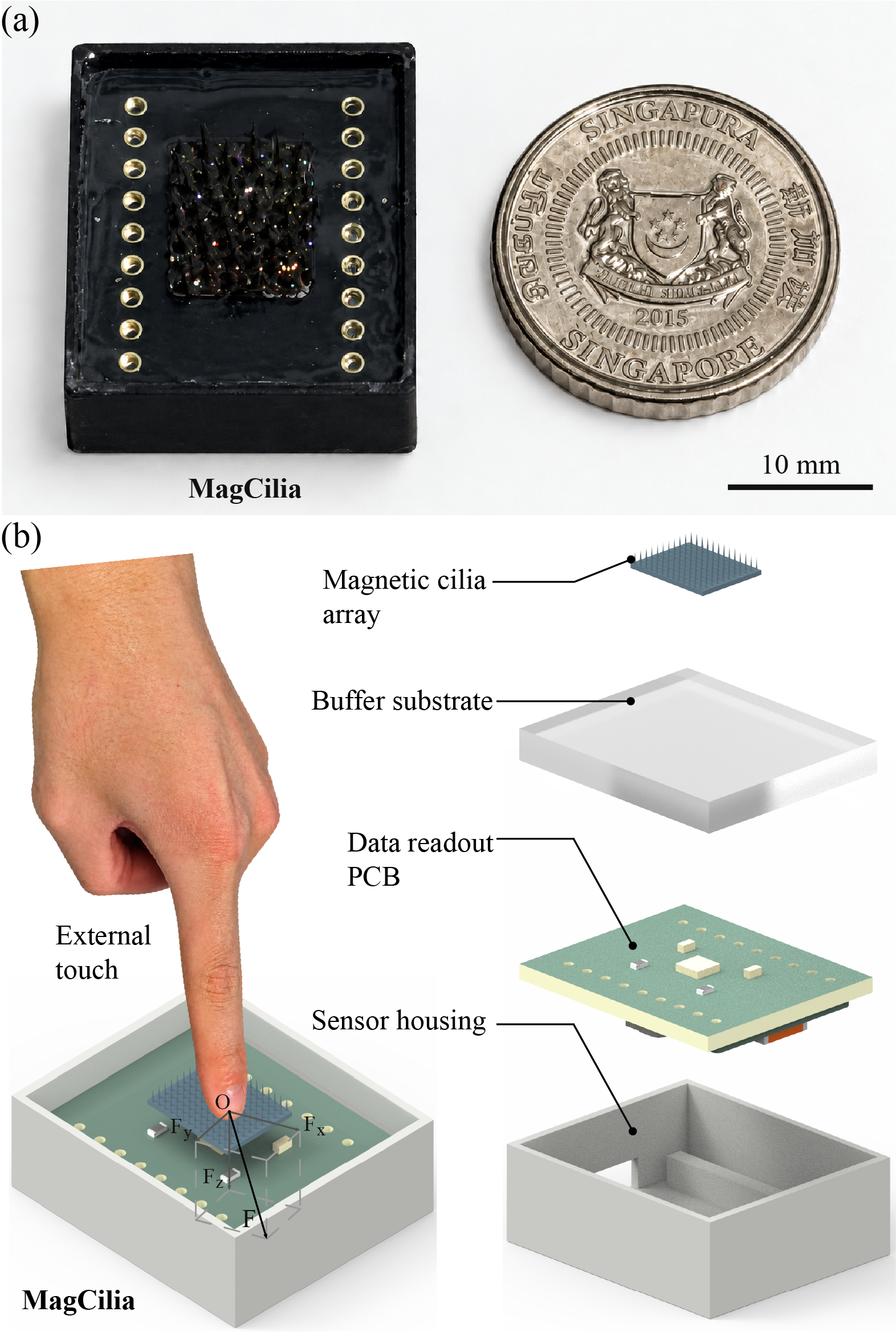}
\caption{Photograph and layered structure of MagCilia. (a) Sensor prototype with a scale reference; (b) magnetic-cilia skin, buffer layer, readout circuit board, and housing.}
\label{fig:structure}
\end{figure}

In recent years, tactile sensors capable of measuring 3D contact forces have provided richer tactile information for dexterous robotic manipulation by sensing both the direction and magnitude of contact loads \cite{Mo2026highly}. Various approaches have been developed for tactile sensing, including piezoresistive \cite{Chen2021flexible}, capacitive \cite{Feng2026intrinsic,Xiang2025highly}, mechanoluminescent imaging \cite{Feng2026soft}, fiber-optic \cite{Li2025fingertip}, visuotactile \cite{Wang2025flexible,Cui2022hand}, and magnetic sensing \cite{Fang2023soft,Liu2026inverse}. Magnetic tactile sensors measure changes in the local magnetic field caused by deformation of flexible magnetic structures under contact loads. Their multi-axis sensing characteristics are closely related to the coupling among the magnetization configuration, structural deformation, and magnetic-field readout \cite{Man2022recent,Peng2025flexible}.

The design of flexible magnetic structures is a key aspect of multidimensional sensing in magnetic tactile sensors. Existing studies have explored various configurations, including sinusoidally magnetized films \cite{Yan2021soft}, stacked orthogonally magnetized films \cite{Yan2024soft}, combinations of magnetic films and flexible pillars \cite{Li2024tactile}, and magnetic cilia \cite{Man2024magnetic,Li2025bioinspired,Liu2026modeling,Feng2026bioinspired,Ji2026design}. Magnetic cilia convert contact loads into changes in the position and orientation of magnetic material through compression and bending, providing a direct mechanical response for sensing contact states and surface features. For magnetic-cilia sensors, quantitative 3D force reconstruction is complicated by coupling between the mechanical deformation and the three magnetic-field channels. In a tactile structure comprising a magnetic-cilia skin and a flexible supporting layer, external loads induce not only compression and bending of the cilia but also displacement of the base membrane and shear deformation of the supporting layer. These deformations jointly alter the local magnetic field, so a force applied along one axis can affect multiple magnetic channels. This coupling limits the use of a direct, channel-by-channel mapping for 3D force reconstruction \cite{Gui2023soft,Zhu2026effect}.

Here, we present MagCilia, a compact single-unit magnetociliary tactile sensor for 3D force sensing. The sensor integrates a magnetic-cilia skin, a PDMS buffer layer, and a Hall sensor. CHFR combines the current three-axis magnetic readout with causal history features to jointly reconstruct the normal force and two tangential force components from the coupled responses. Calibration experiments yield root-mean-square errors of 0.40, 0.57, and 0.69 N for $F_x$, $F_y$, and $F_z$, respectively. Signals are acquired at 246 Hz, and the reconstructed forces are used for robotic grasping feedback and active surface recognition. Robotic experiments demonstrate multi-axis force responses during grasping, stepwise loading, and multidirectional perturbations, with the reconstructed forces used to adjust the gripper opening under external loading. Frequency-domain force features distinguish six surface categories with 99.39\% accuracy in three-fold offline evaluation grouped by acquisition session, and scan-level recognition is demonstrated online. These experiments examine force reconstruction and two robotic applications using the same tactile unit.

The study combines the sensing mechanism, force reconstruction, and robotic use of the reconstructed signals. The grasping experiments examine load magnitude and direction at the contact interface, whereas surface scanning uses the temporal structure of those same force signals to distinguish surface categories.

\section{MagCilia Structure and Sensing Mechanism}
\subsection{Sensor Structure and Fabrication}

MagCilia consists of a magnetic-cilia skin, a polydimethylsiloxane (PDMS) buffer layer, a Hall sensor, and a housing (Fig.~\ref{fig:structure}). The skin contacts the object, and the buffer supports the skin and deforms under load. The housing aligns the layers and readout circuit board. Table~\ref{tab:parameters} summarizes the sensor specifications.

\begin{table}[t]
\caption{Specifications of MagCilia.}
\label{tab:parameters}
\centering
\footnotesize
\begin{tabular}{ll}
\hline
Parameter & Value \\
\hline
Dimensions & $28.00\times24.00\times11.70$ mm \\
Mass & 8.6 g \\
Sampling rate & 246 Hz \\
3D $R^2$ ($F_x/F_y/F_z$) & 0.93 / 0.90 / 0.92 \\
3D RMSE ($F_x/F_y/F_z$) & 0.40 / 0.57 / 0.69 N \\
3D MAE ($F_x/F_y/F_z$) & 0.25 / 0.36 / 0.45 N \\
\hline
\end{tabular}
\end{table}
 
\begin{figure}[htbp]
\centering
\includegraphics[width=0.85\columnwidth]{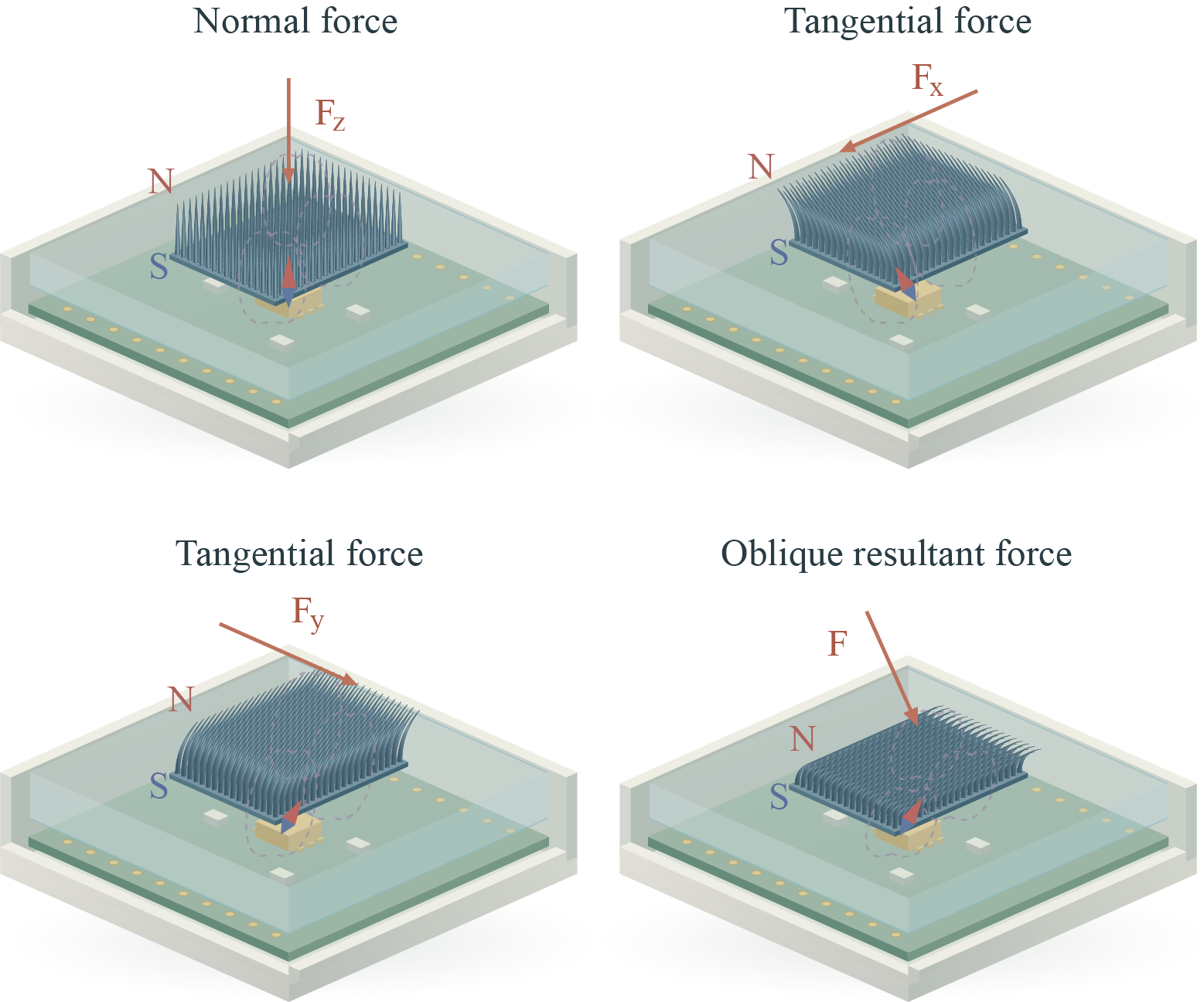}
\caption{Contact sensing mechanism of MagCilia. Normal, $x/y$ tangential, and oblique loading induce different deformations of the magnetic cilia and supporting structure.}
\label{fig:principle}
\end{figure}

The magnetic-cilia skin measures $7.8\times10$ mm, with a 1 mm base membrane and 2.5 mm cilia height. PDMS base, curing agent, and NdFeB powder are mixed at a mass ratio of 10:1:10. An external magnetic field forms the cilia, with magnetization normal to the skin \cite{Feng2026bioinspired}. The $26.15\times22.4\times3.3$ mm buffer uses SYLGARD 184 PDMS, mixed at a base-to-curing-agent ratio of 10:1 and cured at $80\,^{\circ}\mathrm{C}$ for 8 h. PDMS curing bonds the skin to the buffer. The Hall sensor is centered beneath the skin, and signals are acquired at 246 Hz.
 
\subsection{Magnetic Vector Response}
Fig.~\ref{fig:principle} illustrates the typical deformation of MagCilia under normal, tangential, and oblique loading. Normal loading primarily compresses the magnetic cilia and supporting layer, changing the relative position between the magnetic material and the Hall sensing point. Tangential loading bends the cilia and is accompanied by displacement of the magnetic base membrane and shear deformation of the supporting layer, thereby changing both the position and local orientation of the magnetic material. Oblique loading combines normal and tangential components, producing compression, bending, and shear that jointly affect the magnetic-field output. 

The magnetic base membrane and cilia occupy the unloaded reference domain $\Omega_0$. A material point $\mathbf X$ moves by $\mathbf u(\mathbf X)$ under loading. The vector from this deformed point to the Hall sensing point $\mathbf r_h$ is
\begin{equation}
\mathbf r(\mathbf X)=\mathbf r_h-\bigl[\mathbf X+\mathbf u(\mathbf X)\bigr].
\label{eq:v14:2}
\end{equation}
The initial magnetization $\mathbf M_0$ is normal to the skin. Let $\mathbf Q(\mathbf X)$ denote the local rotation tensor from the polar decomposition of the deformation gradient. Conservation of magnetic moment per reference volume gives $\mathrm d\mathbf m=\mathbf Q(\mathbf X)\mathbf M_0\mathrm dV_0$. With relative permeability $\mu_r=1$, superposition of the magnetic dipole fields yields \cite{likharevClassicalElectrodynamics2018}
\begin{equation}
\begin{aligned}
\mathbf B_h={}&\frac{\mu_0}{4\pi}\int_{\Omega_0}
\left[
\frac{3\mathbf r\bigl(\mathbf r^{\mathsf T}\mathbf Q\mathbf M_0\bigr)}{\|\mathbf r\|^5}
-\frac{\mathbf Q\mathbf M_0}{\|\mathbf r\|^3}
\right]\,\mathrm dV_0.
\end{aligned}
\label{eq:v14:3}
\end{equation}
Here, $\mu_0$ is the vacuum permeability, and $\mathbf B_h$ is the three-axis magnetic-field vector at the Hall sensing point. Cilia bending, base-membrane displacement, and buffer-layer deformation change the field through the position and rotation of the magnetic material.

For a fixed contact condition and in the vicinity of a quasi-static loading state $\mathbf{F}^{*}$, a local expansion of the magnetic response gives
\begin{equation}
\begin{aligned}
\delta\mathbf B_h&\simeq\mathbf S(\mathbf F^*)\,\delta\mathbf F, \\
S_{ij}(\mathbf F^*)&=\left.\frac{\partial B_{h,i}}{\partial F_j}\right|_{\mathbf F^*}, \\
&i,j\in\{x,y,z\}.
\end{aligned}
\label{eq:v14:4}
\end{equation}

$\mathbf{S}$ is the local magnetic-response sensitivity matrix. Its off-diagonal terms describe how loading along one direction influences the other magnetic channels. Structural deformation changes the position and orientation of the magnetic material relative to the Hall sensor, producing coupled responses that vary with the loading state \cite{Gui2023soft}. CHFR jointly uses the three magnetic channels and their recent changes to reconstruct forces. Directional information is distributed across the coupled channels and is therefore interpreted jointly.

\subsection{Finite Element Analysis}
Quasi-static finite element analysis in COMSOL examines normal, orthogonal tangential, and oblique loading. The model uses the skin and buffer dimensions above and a regular $9\times13$ cilia array. Cilia root and tip radii are 0.15 and 0.06 mm, with $x/y$ spacings of 0.84/0.78 mm. The magnetic and buffer layers use nearly incompressible neo-Hookean models with shear moduli of 0.31 and 0.5 MPa, respectively, and a bulk-to-shear modulus ratio of 100. The bottom is fixed, and a rigid top plate loads the skin with a contact friction coefficient of 1. A $5\,\mu\mathrm{m}$ normal pre-indentation is maintained while tangential displacements up to $2.5\,\mu\mathrm{m}$ are applied along $\pm x$, $\pm y$, $45^{\circ}$, and $135^{\circ}$. The Hall evaluation point is 3.3 mm below the center of the base membrane.

\begin{figure}[htbp]
\centering
\includegraphics[width=0.85\columnwidth]{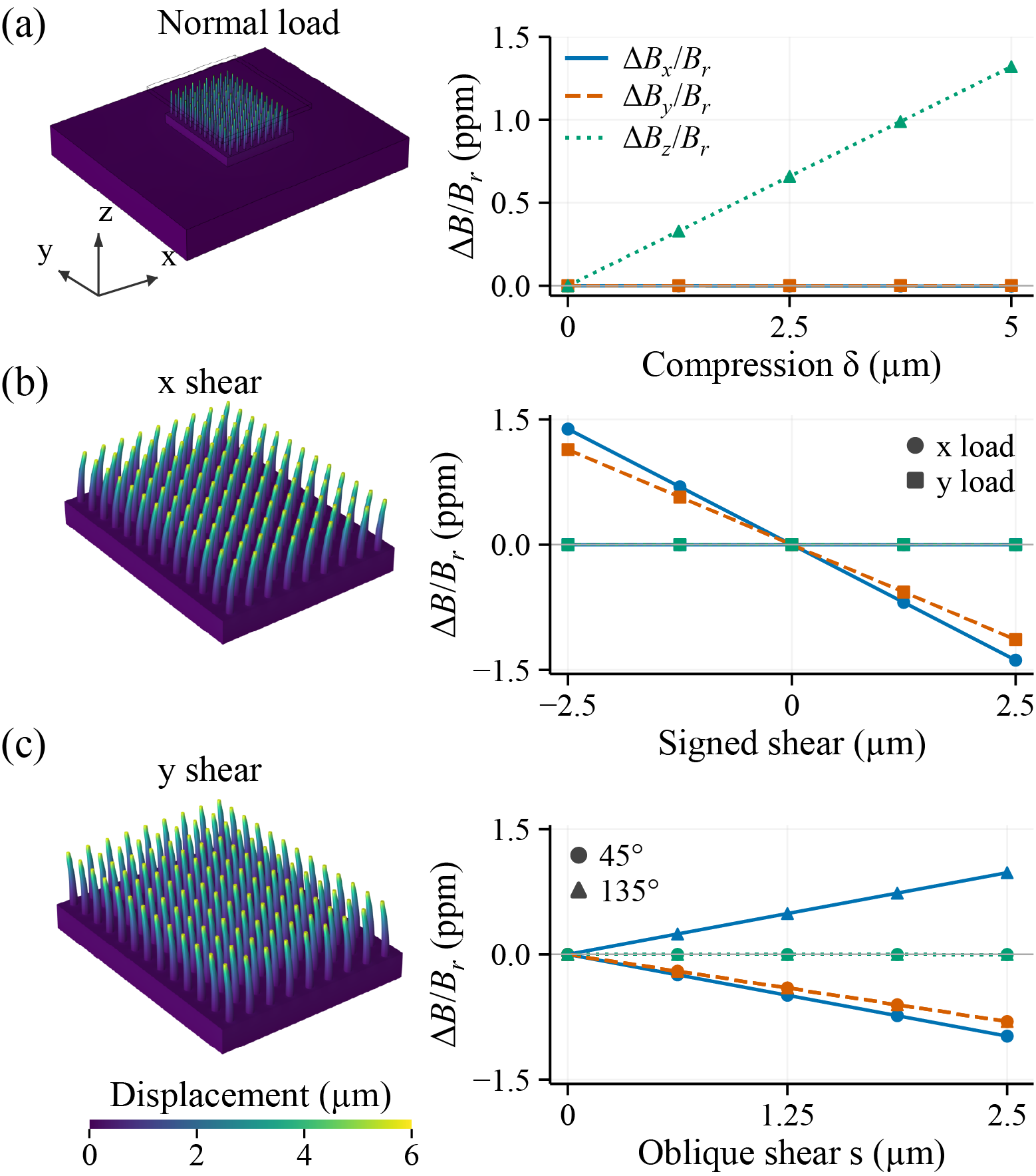}
\caption{Quasi-static deformation and normalized magnetic response. (a) Normal indentation and corresponding magnetic response; (b) $x$ shear deformation (left) and magnetic response under orthogonal shear (right); (c) $y$ shear deformation (left) and magnetic response under oblique shear (right).}
\label{fig:simulation}
\end{figure}

The magnetic-field changes are normalized by the remanence parameter $B_r$ and expressed as $10^6\Delta B/B_r$. The unloaded state is used as the baseline for normal loading, whereas the pre-indented state is used for tangential loading. Fig.~\ref{fig:simulation} shows the magnetic-field changes under different loading directions. Under normal indentation, the magnetic-field change occurs primarily in the $B_z$ component. Under shear along the positive and negative orthogonal directions, the dominant transverse magnetic-field response reverses sign with the loading direction, whereas oblique shear produces changes in both transverse components. Different loading directions therefore produce distinct, coupled magnetic responses. These directional responses motivate calibration using the complete magnetic-field vector.
 
\section{3D Force Calibration and Reconstruction}
\subsection{Synchronized Data Acquisition and Organization}
\begin{figure}[htbp]
\centering
\includegraphics[width=\columnwidth]{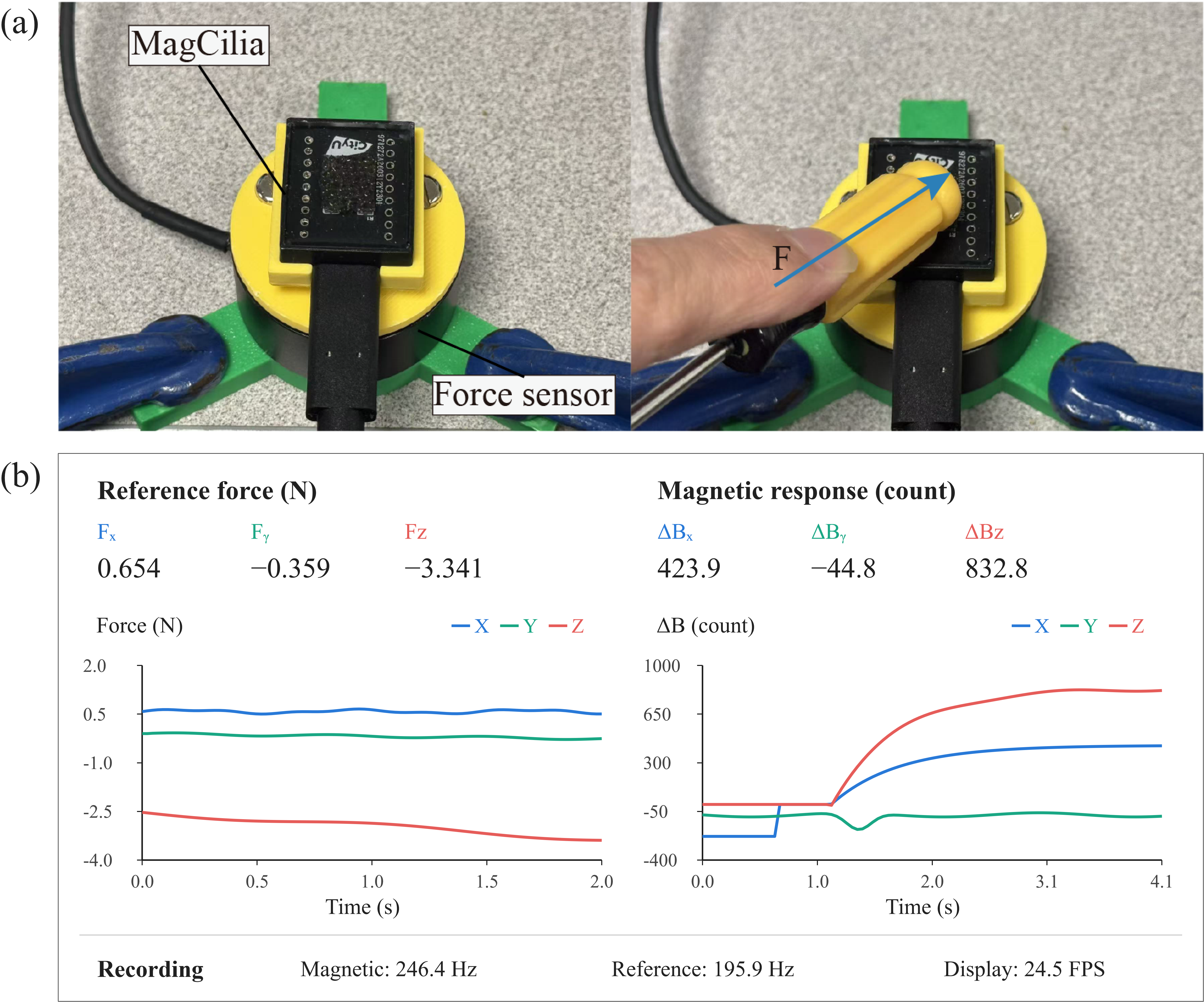}
\caption{3D force calibration setup and data acquisition interface. (a) Sensor installation and contact loading; (b) 3D reference force, magnetic readout, and synchronized recording.}
\label{fig:calibration}
\end{figure}

The calibration setup consists of MagCilia and a reference force sensor (Dayang Sensor Y45), as shown in Fig.~\ref{fig:calibration}. To calibrate the coupled magnetic response across different contact states, a handheld indenter applies normal compression, lateral loading, and combinations of both, with the direction and magnitude varied during each record. The 3D reference force and magnetic-field signals are acquired synchronously. Five independent loading records are collected, comprising 304,870 synchronized magnetic-field and reference-force sample pairs. The data cover force ranges of $-4.74$--$4.81$ N for $F_x$, $-6.64$--$4.74$ N for $F_y$, and $-9.60$--$0.05$ N for $F_z$. Here, $F_x$ and $F_y$ are tangential components, and $F_z$ is normal to the sensing surface. Compression corresponds to negative $F_z$, with $F_n=-F_z$ denoting positive normal compression in the grasping experiments.
 
\subsection{Causal History Features}
To represent the loading history of the flexible structure, CHFR augments the baseline-corrected three-axis magnetic readout $\Delta\mathbf{s}(t)$ with differences over 40 and 120 ms. For an interval $\tau$, we define

\begin{equation}
\mathbf d_{\tau}(t)=\Delta\mathbf s(t)-\Delta\mathbf s(t-\tau),
\label{eq:v14:5}
\end{equation}
and construct the nine-dimensional input
\begin{equation}
\begin{aligned}
\mathbf x(t)={}&\bigl[\Delta\mathbf s(t)^{\mathsf T},\mathbf d_{40\mathrm{ms}}(t)^{\mathsf T},\;\mathbf d_{120\mathrm{ms}}(t)^{\mathsf T}\bigr]^{\mathsf T}.
\end{aligned}
\label{eq:v14:6}
\end{equation}

Historical samples are selected by timestamp using the latest available measurement at or before $t-\tau$. Each vector contains all three magnetic channels. The current readout and its recent differences form a causal input that is computed continuously during acquisition. The readout represents the current magnetic state, while the differences describe its recent evolution at two time scales. These features retain cross-channel information and represent changes in the magnetic response.
 
\subsection{3D Force Reconstruction}
\begin{figure}[htbp]
\centering
\includegraphics[width=0.85\columnwidth]{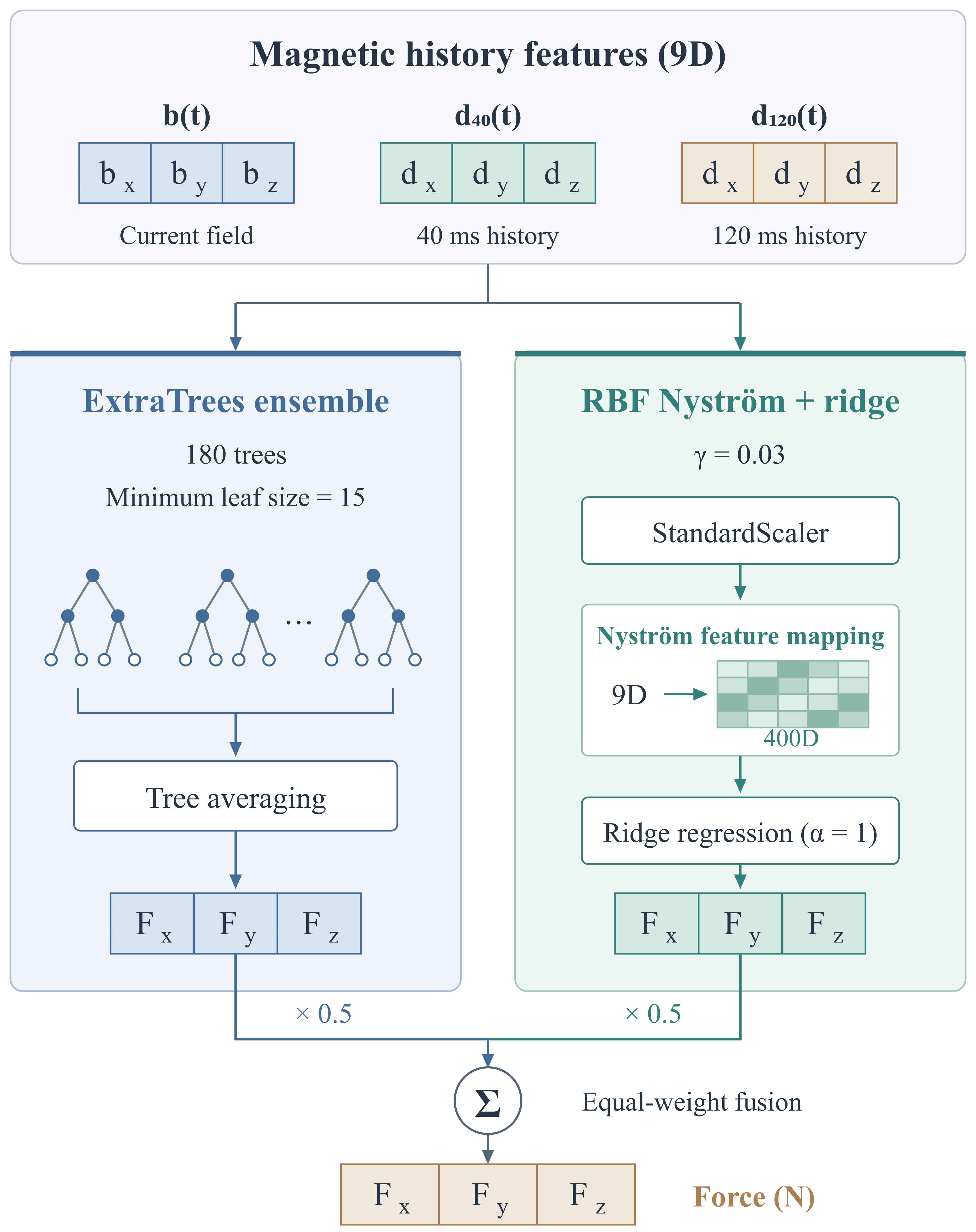}
\caption{Causal history fusion regression (CHFR) model for 3D force reconstruction.}
\label{fig:force-model}
\end{figure}

CHFR maps the nine-dimensional causal history features to 3D contact forces using two parallel branches (Fig.~\ref{fig:force-model}): Extremely Randomized Trees (Extra Trees, ET) \cite{geurtsExtremelyRandomizedTrees2006} and kernel-approximation ridge regression. The ET branch consists of 180 decision trees, with at least 15 samples per leaf, to learn the nonlinear mapping between the magnetic features and contact forces. In the kernel-regression branch, the input features are first standardized using StandardScaler, followed by an approximation of the radial basis function (RBF) kernel using 400 Nystr\"om components \cite{williamsNystromKernelMachines2001}, with kernel parameter $\gamma=0.03$ and ridge regularization parameter $\alpha=1$. StandardScaler and the Nystr\"om map are fitted on each fold's training data. Each branch predicts $F_x$, $F_y$, and $F_z$, and the predictions are averaged to obtain the final force estimate:
\begin{equation}
\widehat{\mathbf F}(t)=\frac12\widehat{\mathbf F}_{\mathrm{ET}}(t)+\frac12\widehat{\mathbf F}_{\mathrm{RBF}}(t).
\label{eq:v14:7}
\end{equation}
The ET branch captures nonlinear relationships through local partitions, whereas the kernel branch provides a smooth nonlinear approximation. Both branches use the same inputs and predict forces in the same mechanical coordinate system.
 
Five-fold cross-validation holds out one complete calibration record per fold and trains on the remaining four. Up to 100,000 training points are randomly sampled for ET and 70,000 for kernel regression, using a fixed seed. Each test record is retained in full. Model selection and evaluation use the same five record-level folds. Performance is calculated from pooled out-of-fold predictions using the coefficient of determination $R^2$, root-mean-square error (RMSE), and mean absolute error (MAE). RMSE and MAE express force errors in newtons, while $R^2$ measures agreement relative to the variation of the reference force.

\section{3D Force Sensing Performance}
\subsection{3D Force Reconstruction}
\begin{figure}[htbp]
\centering
\includegraphics[width=\columnwidth]{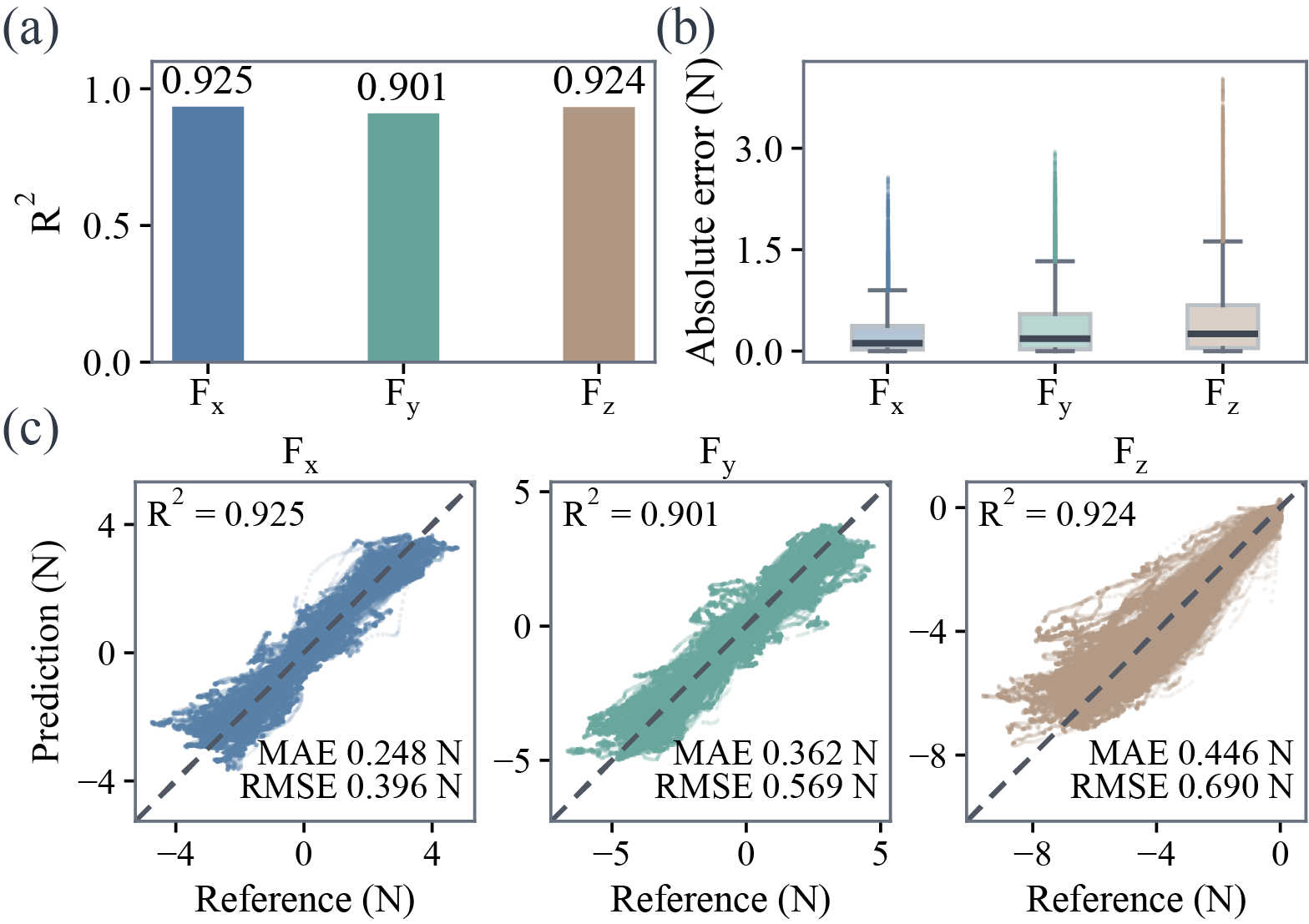}
\caption{3D force reconstruction results. (a) Coefficient of determination for each force component; (b) absolute-error distributions; (c) scatter plots of reference forces versus out-of-fold predictions with error metrics.}
\label{fig:force-accuracy}
\end{figure}

Fig.~\ref{fig:force-accuracy} summarizes reconstruction performance using the pooled out-of-fold predictions from the five held-out calibration records. The $R^2$ values for $F_x$, $F_y$, and $F_z$ are 0.93, 0.90, and 0.92, respectively, with corresponding RMSE values of 0.40, 0.57, and 0.69 N and MAE values of 0.25, 0.36, and 0.45 N. These results support joint reconstruction of normal and tangential forces within the calibrated load ranges and contact conditions. The dynamic-touch experiments below complement the aggregate metrics by examining temporal force tracking.

\begin{figure*}[t]
\centering
\includegraphics[width=0.94\textwidth]{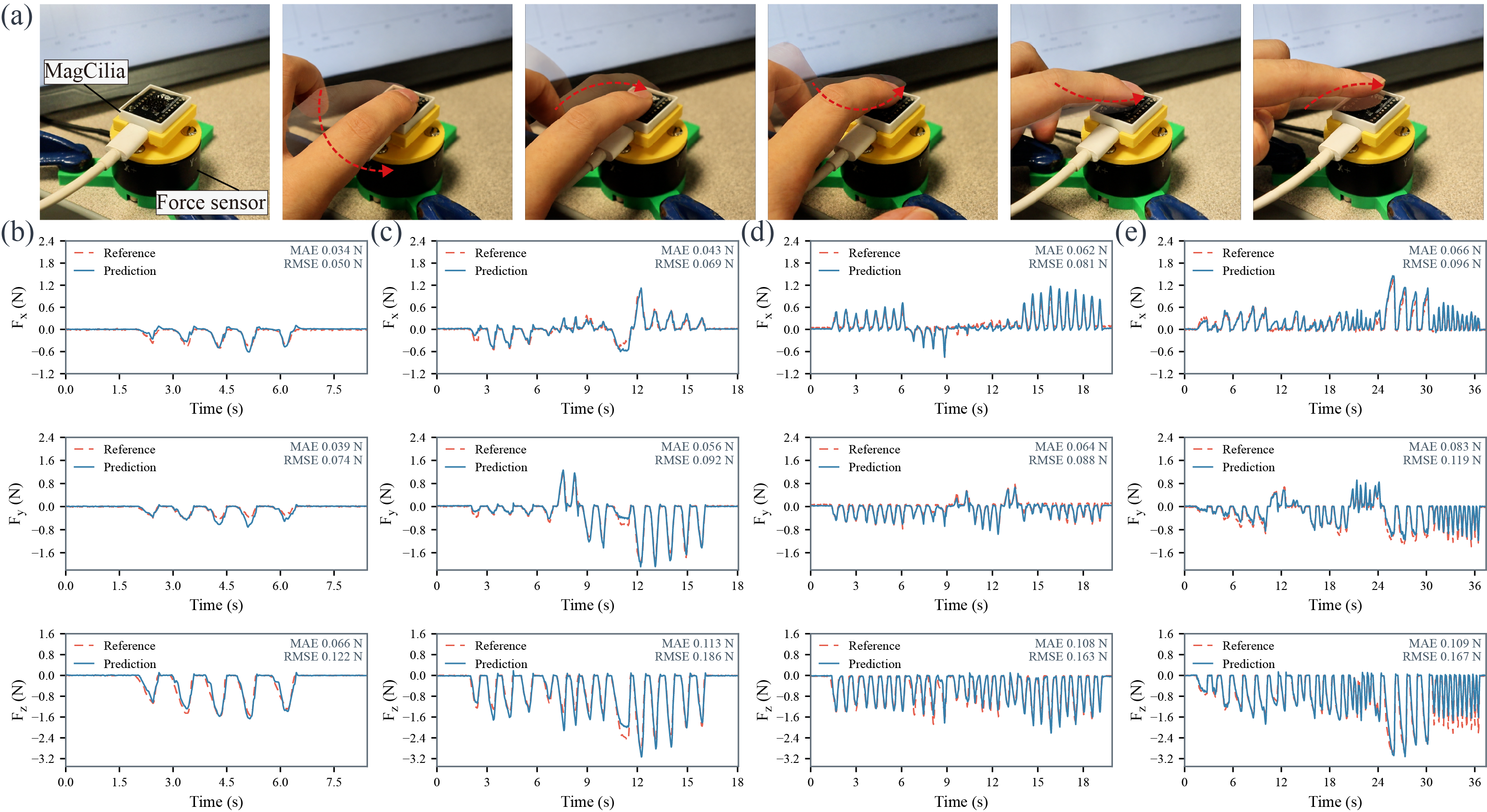}
\caption{Dynamic-touch response. (a) Fingertip loading and reference-sensor setup; (b-e) reference forces (red dashed) and predictions (blue solid) during contacts with varying magnitudes, directions, and repetition rates.}
\label{fig:dynamic-touch}
\end{figure*}

\subsection{Dynamic Touch Response}
To examine tracking under changing contact loads, MagCilia is mounted on the reference force sensor and loaded manually with a fingertip (Fig.~\ref{fig:dynamic-touch}(a)). Repeated pressing, release, lateral fingertip loading, and combined loading vary contact magnitude, direction, and repetition rate, including rapid successive contacts. Contact strength varies between repetitions, so the records examine changes in force amplitude as well as direction and repetition rate. Synchronized reference forces and CHFR predictions are compared in Fig.~\ref{fig:dynamic-touch}(b-e). The reconstructed normal force follows contact, compression, and unloading; tangential forces follow changes in interaction direction. In Fig.~\ref{fig:dynamic-touch}(b), RMSEs are 0.05, 0.07, and 0.12 N for $F_x$, $F_y$, and $F_z$, respectively.
 
During combined loading, normal pressing and lateral motion act together, and all three predictions track the corresponding reference-force variations. Across valid paired time points in the four records, RMSEs range from 0.05 to 0.10 N for $F_x$, 0.07 to 0.12 N for $F_y$, and 0.12 to 0.19 N for $F_z$.

\section{Demonstration on Robotic Grasping}
\subsection{CHFR-Based Force Feedback and Gripper Adjustment}
\begin{figure*}[t]
\centering
\includegraphics[width=0.94\textwidth]{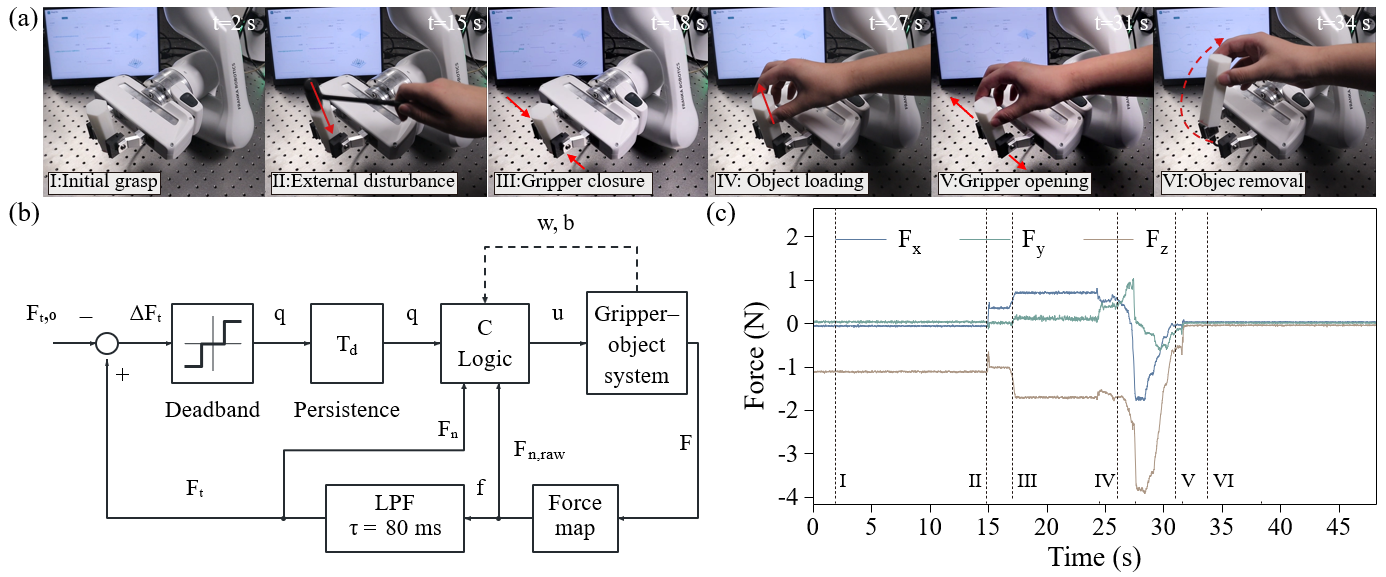}
\caption{Force-guided robotic grasping with MagCilia. (a) Grasping, perturbation, loading, and object-removal stages; (b) force-feedback and state logic for adjusting the gripper opening; (c) corresponding CHFR-reconstructed 3D forces.}
\label{fig:grasp-control}
\end{figure*}

MagCilia is mounted on the gripper of a robot (Franka Research 3). CHFR reconstructs $F_x$, $F_y$, and $F_z$ from the magnetic readout in real time and supplies these forces to the grasping controller. The experiment tests whether tangential-force feedback can support both grasp retention and handover. After a stable grasp is established, downward disturbances and additional loading are applied to emulate a tendency to drop; a person then supports and pulls the object upward for removal (Fig.~\ref{fig:grasp-control}(a)). The controller adjusts the gripper opening according to these load changes using the state logic in Fig.~\ref{fig:grasp-control}(b).

The reconstructed force signals are low-pass filtered with a time constant of 80 ms. Let $F_t$ denote the tangential force component aligned with the applied vertical loading, and let $F_{t,0}$ denote its baseline during stable grasping. The change $\Delta F_t=F_t-F_{t,0}$ passes through a deadband and a persistence criterion before entering the control logic. The logic issues closing, opening, or holding commands according to the load direction, normal force, and gripper state. Sustained downward loading triggers additional closure to strengthen the grasp, whereas sustained upward support or pulling triggers opening so that the person can take the object. When the tangential force returns to the baseline region, the gripper holds its current opening. The normal compressive force is defined as $F_n=-F_z$. Its raw and filtered values are used to monitor contact and limit further closure during adjustment. Tangential-force changes select the adjustment direction, while normal-force feedback constrains contact compression during closure.

Fig.~\ref{fig:grasp-control}(c) shows the reconstructed forces throughout the interaction. During the initial grasping stage from 2 to 14 s, $F_n$ remains at $\sim 1.1$ N and the tangential forces remain close to zero. With additional external loading, the controller tightens the grasp, and $F_n$ increases to $\sim 1.7$ N during 18--24 s while $F_x$ increases to $\sim 0.7$ N. Further loading produces a peak $F_n$ of 3.91 N, accompanied by changes in the tangential forces. During upward support and object removal, the controller opens the gripper, and all three force components return to near zero after contact is released. The sequence demonstrates force-guided adjustment during loading and handover: increased downward loading leads to greater contact compression, whereas upward support allows the grasp to be released for object removal.

\subsection{Stepwise Loading During Grasping}
\label{sec:stepwise}
\begin{figure}[t]
\centering
\includegraphics[width=0.9\columnwidth]{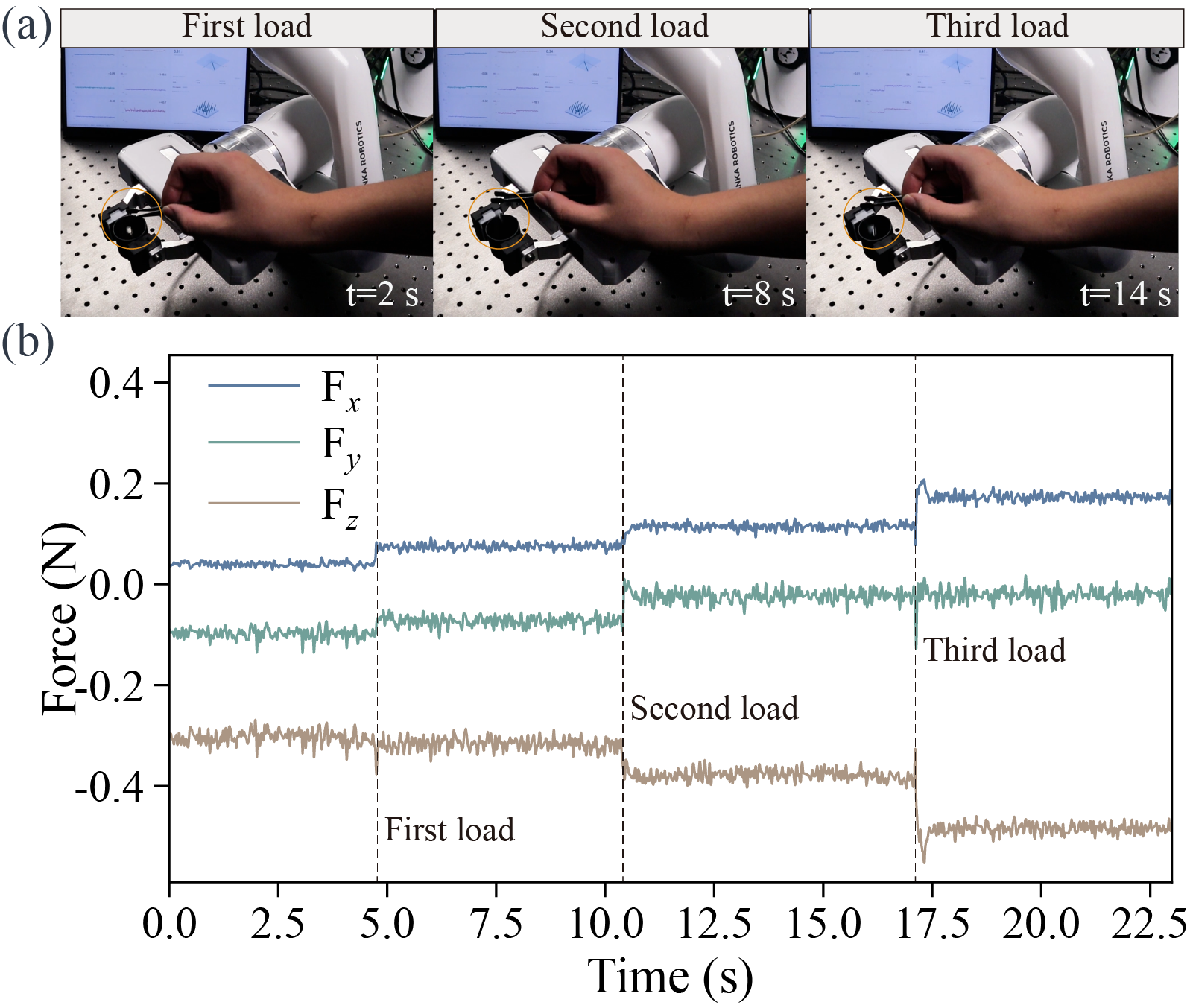}
\caption{Stepwise loading during grasping. (a) Three successive weight additions to the box; (b) CHFR-reconstructed 3D forces, with settled $F_x$ plateaus indicating cumulative tangential loading.}
\label{fig:grasp-loading}
\end{figure}

To examine the response to incremental tangential loading, weights are added to a grasped box in three successive steps, allowing the force response to settle after each addition (Fig.~\ref{fig:grasp-loading}(a)). Each weight remains in the box when the next is added, creating a cumulative loading sequence. The settled $F_x$ plateaus are compared in Fig.~\ref{fig:grasp-loading}(b), separating the persistent load increments from transient changes during weight addition. Before loading, the mean reconstructed $F_x$ is 0.04 N. The mean $F_x$ values at the three stable plateaus increase to 0.08, 0.11, and 0.17 N, respectively. Meanwhile, $F_n$ increases from 0.32 to 0.48 N, while $F_y$ remains within a relatively small range. These distinct $F_x$ plateaus demonstrate the response to small load increments transmitted through the grasping interface, complementing the larger load changes in the gripper-adjustment experiment.

\subsection{External Perturbations in Different Directions}
\begin{figure}[htbp]
\centering
\includegraphics[width=0.9\columnwidth]{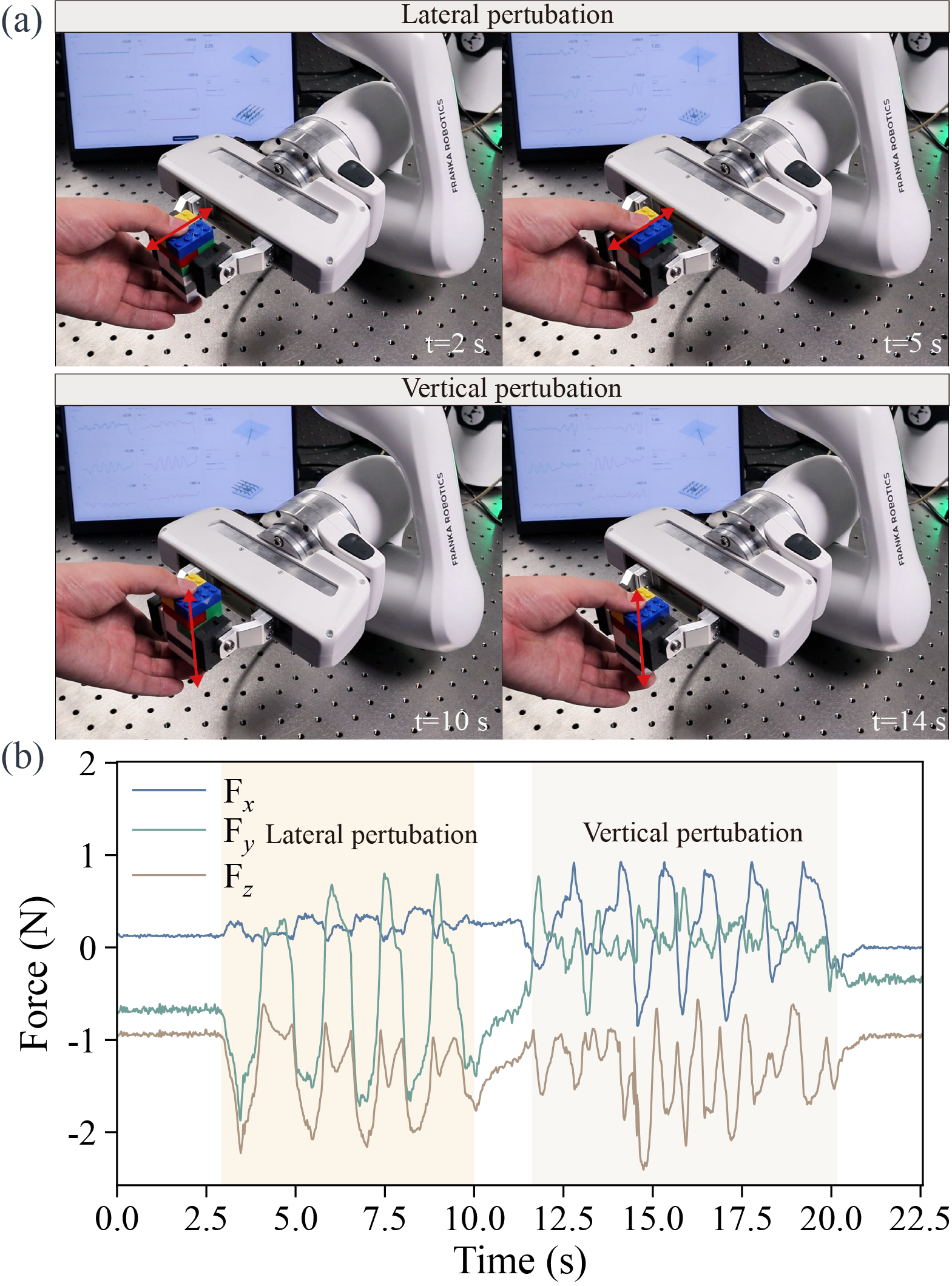}
\caption{Lateral and vertical perturbations during grasping. (a) Manual perturbation actions; (b) simultaneous CHFR-reconstructed 3D forces, with shaded intervals identifying lateral and vertical perturbations.}
\label{fig:grasp-perturbation}
\end{figure}

To examine directional load changes at the grasping interface, the object is perturbed laterally and vertically after a grasp is established, while all three force components are recorded (Fig.~\ref{fig:grasp-perturbation}). These actions test whether the force components distinguish changing interaction directions while the object remains grasped. During lateral perturbation, $F_y$ varies bidirectionally over a range of $-1.87$--$0.80$ N, accompanied by changes in $F_x$ and $F_n$. Changing the perturbation direction produces different magnitudes and temporal responses across the three force components. During vertical perturbation, $F_x$ and $F_y$ span $-0.85$--$0.93$ N and $-0.73$--$0.77$ N, respectively, while $F_n$ reaches a peak of 2.41 N. The concurrent changes in normal and tangential forces characterize the combined interface loading caused by these perturbations.

\begin{figure*}[t]
\centering
\includegraphics[width=0.95\textwidth]{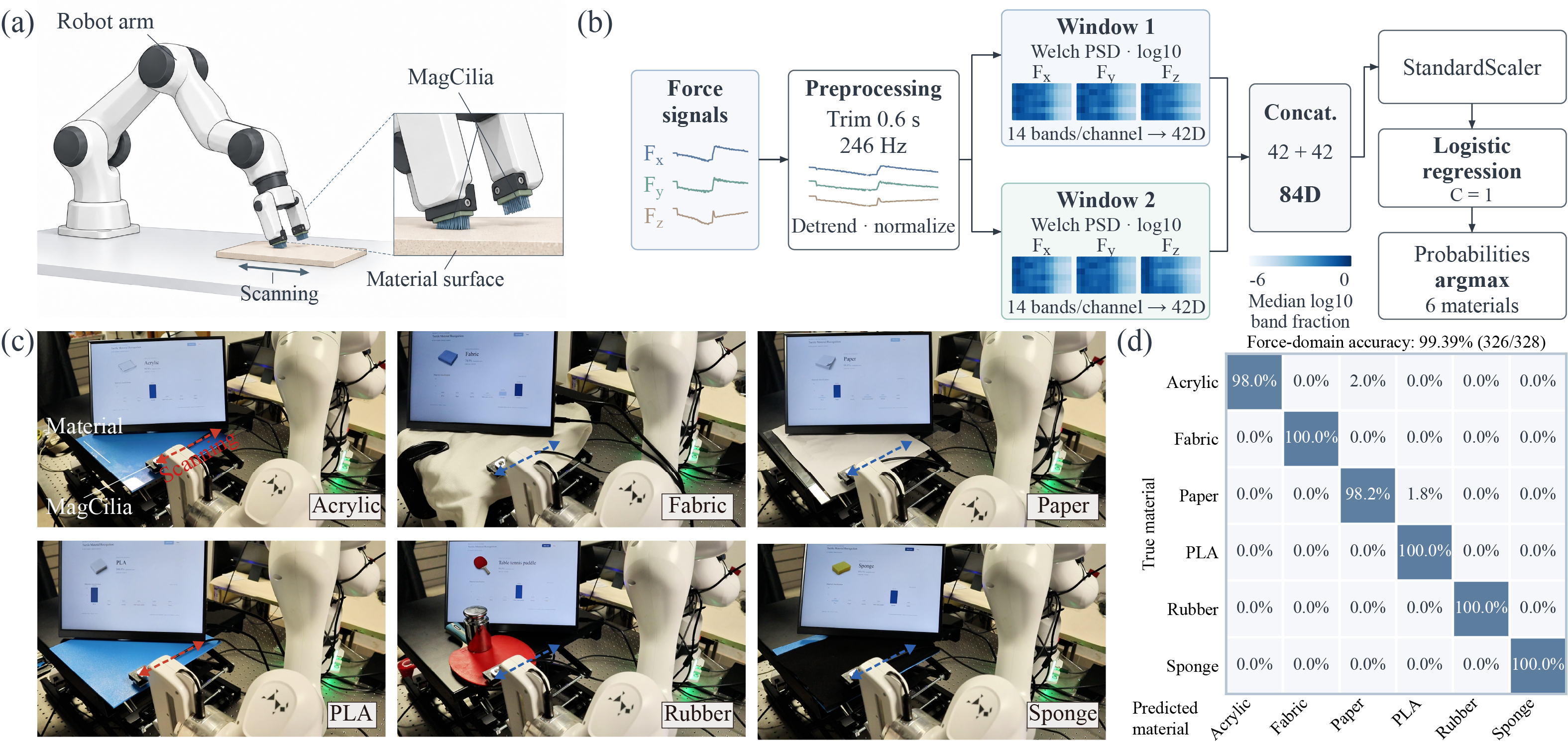}
\caption{Robotic surface scanning and recognition. (a) Robotic scanning setup with MagCilia; (b) 3D force frequency-band features and classification pipeline; (c) real-time robotic recognition demonstration for six surface categories; (d) row-normalized confusion matrix from three-fold session-grouped offline evaluation, with an overall accuracy of 99.39\%.}
\label{fig:material-classification}
\end{figure*}

\section{Demonstration on Surface Recognition}
\subsection{Active Scanning and Sample Composition}
Active contact provides surface-dependent tactile signals for surface recognition \cite{Qin2024surface}. Fig.~\ref{fig:material-classification}(a) shows the robotic scanning setup. MagCilia moves forward and returns along each surface while maintaining cilia contact. Each forward-and-return scan forms one trial, with the three force components recorded throughout both directions. This procedure examines surface-dependent contact fluctuations during scanning. The dataset contains 328 trials across 52 acquisition sessions, covering acrylic, fabric, paper, polylactic acid (PLA), rubber, and sponge. Fig.~\ref{fig:material-classification}(b) shows the frequency-feature and logistic-regression pipeline.

\begin{figure}[t]
\centering
\includegraphics[width=0.9\columnwidth]{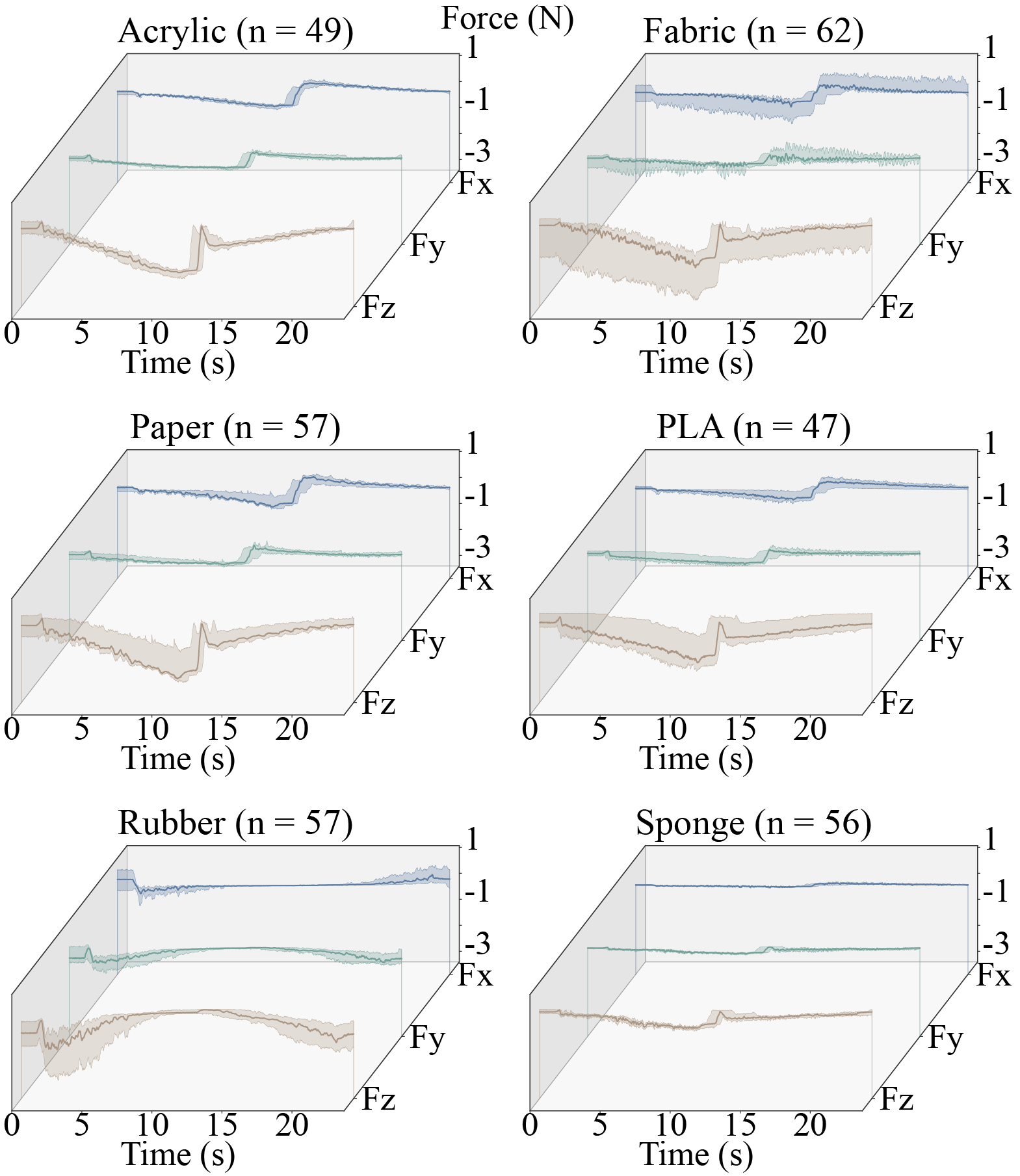}
\caption{CHFR-reconstructed 3D force time series during scanning of six surface categories. Solid lines show pointwise medians across trials; shaded regions show the minimum--maximum range. $n$ denotes the number of trials.}
\label{fig:material-signals}
\end{figure}

The six surface categories produce distinct force amplitudes and temporal fluctuations (Fig.~\ref{fig:material-signals}). The peak magnitude of the median normal-force curve is 2.50 N for paper and 0.73 N for sponge. The median $F_x$ curves have peak-to-peak amplitudes of 1.18 and 0.18 N, respectively. These multi-axis scanning responses supply the signals used for frequency-domain surface recognition. The median traces summarize the typical response, while the shaded ranges in Fig.~\ref{fig:material-signals} show variation among repeated trials.

\subsection{Frequency-Domain Representation}
Scanning signals are resampled at 250 Hz and detrended before spectral analysis. Welch's method estimates the power spectrum of each force component in each scanning direction \cite{welchPowerSpectra1967}. Spectra are divided into 14 frequency bands. For direction $d$, component $j$, and band $I_k$, band power and its normalized logarithmic feature are
\begin{equation}
\begin{aligned}
E_{djk}&=\sum_{f\in I_k}P_{dj}(f), \\
z_{djk}&=\log_{10}\!\left(\max\!\left[\frac{E_{djk}}{\sum_{\ell=1}^{14}E_{dj\ell}},\epsilon\right]\right),
\end{aligned}
\label{eq:frequency-features}
\end{equation}
where $P_{dj}(f)$ is the power spectrum and $\epsilon>0$ sets a lower bound on normalized band power. Concatenating two directions, three force components, and 14 bands gives 84 features. Within-channel normalization represents relative spectral distributions and reduces dependence on overall force amplitude.

The two scanning directions are retained separately so that direction-dependent force fluctuations contribute to recognition. The classifier therefore uses the distribution of relative band power across both tangential and normal force components, complementing the time-domain responses in Fig.~\ref{fig:material-signals}.

\subsection{Classification Results}
In the online demonstration (Fig.~\ref{fig:material-classification}(c)), each forward-and-return scan is converted into frequency features and classified. All six tested surfaces were correctly identified. Offline evaluation uses three-fold cross-validation grouped by acquisition session (Fig.~\ref{fig:material-classification}(d)). All trials from one session are assigned to the same fold. StandardScaler is fitted on each training set and applied to the held-out sessions, followed by multiclass logistic regression. The pooled out-of-fold predictions correctly classify 326 of 328 trials, yielding 99.39\% accuracy, 99.37\% balanced accuracy, and a macro-averaged F1 score of 0.99. These results demonstrate surface-category discrimination across held-out acquisition sessions.

\section{Comparison With Related Work}
Magnetic multi-axis tactile sensors typically combine magnetic-structure design with signal reconstruction to map contact-induced structural deformation to force information along different directions (Table~\ref{tab:sensor-comparison}). Yan et al. used a sinusoidally magnetized film to achieve self-decoupled sensing of the normal force and one tangential force component \cite{Yan2021soft}, and further employed two orthogonally magnetized film layers to extend the measurement of a single tactile unit to three force components \cite{Yan2024soft}. Another class of studies exploits the bending and orientation changes of magnetic cilia to acquire contact information. Man et al. combined magnetic-cilia arrays with magnetic-sensor arrays to achieve sliding sensing and object recognition through spatiotemporal magnetic responses \cite{Man2024magnetic}. Li et al. used the giant magnetoimpedance effect of amorphous soft-magnetic ribbons to sense weak tangential interactions \cite{Li2025bioinspired}. Liu et al. also developed a segmented-beam model comprising a nonmagnetic base segment and a magnetic tip to describe cilia deformation and magnetic response, and further applied it to sensing flow direction and surface texture \cite{Liu2026modeling}.
 
MagCilia uses a single magnetociliary contact unit and a single Hall sensor to acquire contact-induced changes in the magnetic field, and CHFR simultaneously reconstructs $F_x$, $F_y$, and $F_z$ from the coupled magnetic responses. The reconstructed 3D forces are then used for force feedback during robotic grasping and, through frequency-domain features obtained during active scanning, for surface recognition. Table~\ref{tab:sensor-comparison} compares sensor configurations, sensing capabilities, and quantitative performance reported for the respective sensing tasks. This comparison places the single-unit 3D force reconstruction of MagCilia alongside the configurations and functions of related magnetic tactile sensors.

\begin{table*}[t]
\caption{Comparison of MagCilia with representative magnetic tactile sensors.}
\label{tab:sensor-comparison}
\centering
\footnotesize
\setlength{\tabcolsep}{5pt}
\renewcommand{\arraystretch}{1.15}
\begin{tabularx}{\textwidth}{
l
>{\raggedright\arraybackslash}p{0.32\textwidth}
>{\raggedright\arraybackslash}p{0.25\textwidth}
>{\raggedright\arraybackslash}X}
\hline
Sensor & Configuration & Capability & Quantitative performance \\
\hline

\cite{Yan2021soft}
& Magnetized film with Hall sensing
& Normal and 1D tangential force
& Sensitivity: 0.01 kPa$^{-1}$ (normal), 0.10--0.27 kPa$^{-1}$ (tangential) \\

\cite{Yan2024soft}
& Stacked magnetic films with a $3\times3$ Hall array
& 3D force sensing
& Range: 28 N (normal), $\pm 8$ N (tangential) \\

\cite{Man2024magnetic}
& Magnetic-cilia array with a $2\times4$ sensor array
& Tangential force and direction
& Resolution: 0.2 mN \\

\cite{Li2025bioinspired}
& Soft-magnetic cilia with an induction coil
& Weak tangential force sensing
& Sensitivity: 10.03\%/mN \\

\cite{Liu2026modeling}
& Segmented magnetic cilia
& Flow and texture sensing
& Flow-direction error: $<10^{\circ}$ \\

\cite{Feng2026bioinspired}
& Magnetic cilia with a single Hall sensor
& Normal force and tangential direction
& Normal sensitivity: 0.015 N$^{-1}$; $R^2=0.97$ \\

\textbf{This work}
& \textbf{Magnetic cilia with a single Hall sensor}
& \textbf{3D force sensing}
& \textbf{RMSE: 0.40/0.57/0.69 N;}\newline
  \textbf{$R^2$: 0.93/0.90/0.92} \\
  
\hline
\end{tabularx}
\end{table*}

\section{Conclusion}

Robotic manipulation requires compact tactile sensors capable of resolving multidirectional contact forces while providing rich information about physical interactions. We present MagCilia, a single-unit magnetociliary tactile sensor that combines a flexible magnetic-cilia structure with a single Hall sensor for 3D force sensing. CHFR reconstructs the three force components with RMSEs of 0.40, 0.57, and 0.69 N for $F_x$, $F_y$, and $F_z$, respectively, in grouped five-fold cross-validation. Robotic experiments demonstrate tangential-force-guided gripper adjustment during external loading and handover, together with monitoring of multidirectional perturbations. Frequency-domain force features classify six surface categories with 99.39\% accuracy in three-fold offline cross-validation grouped by acquisition session. The online demonstration correctly identified all six tested surfaces. These results demonstrate how one tactile unit can provide both quantitative contact-force feedback and surface information through the same reconstructed signals. Future work will optimize the magnetic composite and cilia geometry and integrate a Hall-sensor array to enhance force-reconstruction accuracy and spatial resolution, supporting broader robotic applications.

\bibliographystyle{IEEEtran}
\bibliography{references}

\makeatletter
\def\@IEEEBIOskipN{2\baselineskip}
\makeatother

\begin{IEEEbiography}[{\includegraphics[width=1in,height=1.25in,clip,keepaspectratio]{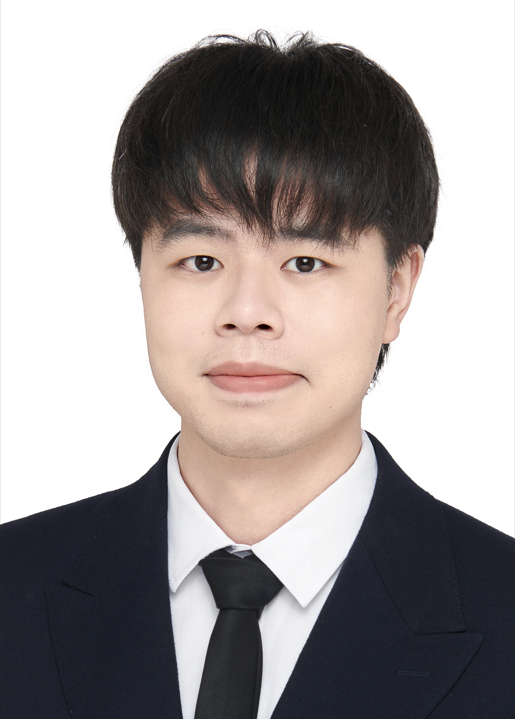}}]{Yu Feng}
(Member, IEEE) received the B.Eng. degree in Measurement and Automation from Harbin Engineering University, Harbin, China, in 2020, and the Ph.D. degree in Mechanical Engineering from City University of Hong Kong, Hong Kong SAR, in 2024. He subsequently worked as a postdoctoral fellow at City University of Hong Kong from 2024 to 2026. He is currently a research fellow with the Department of Mechanical Engineering, National University of Singapore, Singapore. His research interests include robotic perception, tactile sensors, and robot manipulation.
\end{IEEEbiography}

\begin{IEEEbiography}
[{\includegraphics[width=1in,height=1.25in]{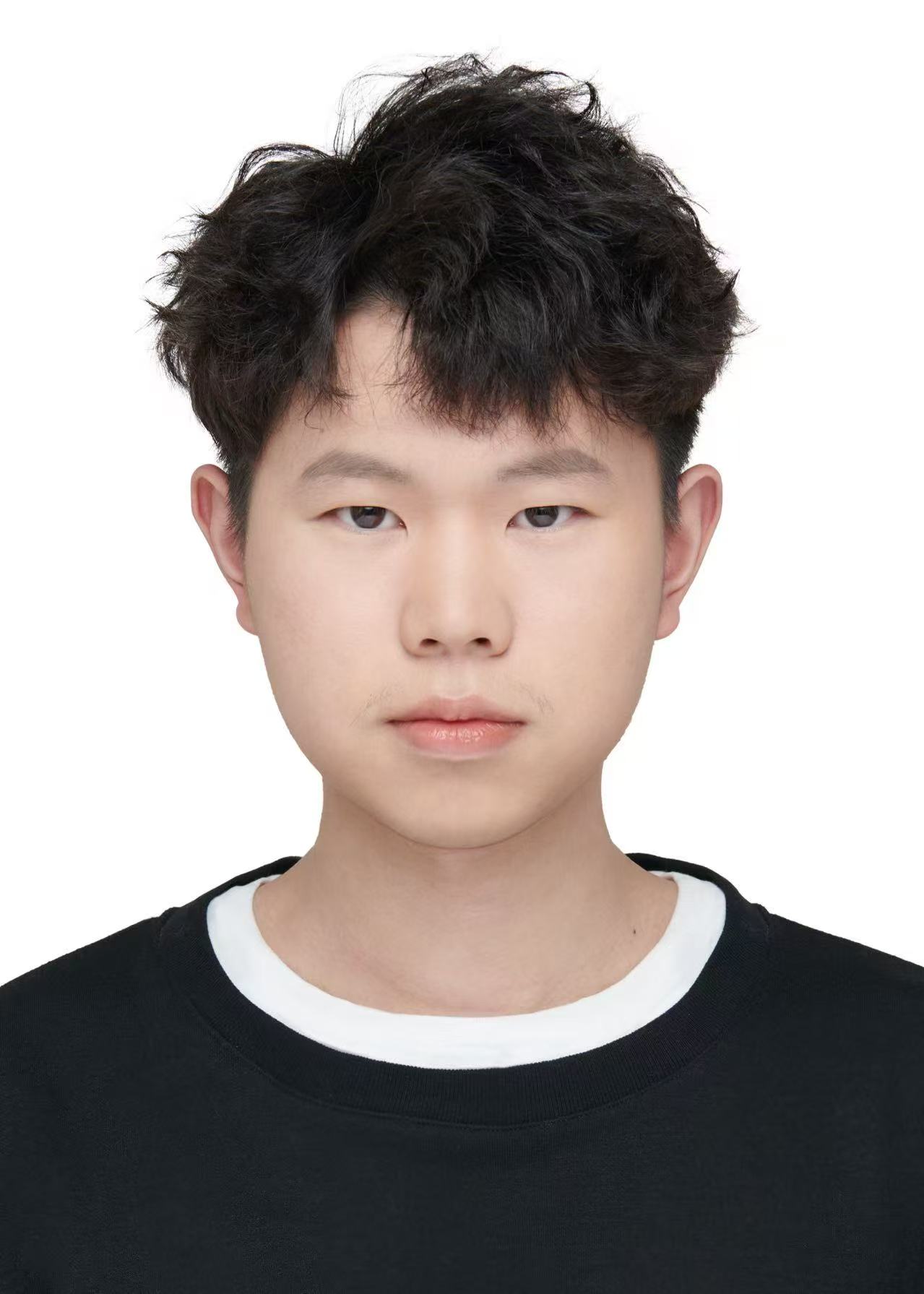}}]
{Hao Wu} (Student Member, IEEE) received the B.E. degree in Mechanical Engineering from Zhejiang University, China, in 2024. From 2024 to 2026, he worked as a research assistant with the Grasp Lab at Zhejiang University. He is currently pursuing a Ph.D. degree at the National University of Singapore, Singapore, under the supervision of Prof. Jianshu Zhou. His research interests include mechanism design, soft robotics, and multimodal tactile perception.
\end{IEEEbiography}

\begin{IEEEbiography}
[{\includegraphics[width=1in,height=1.25in]{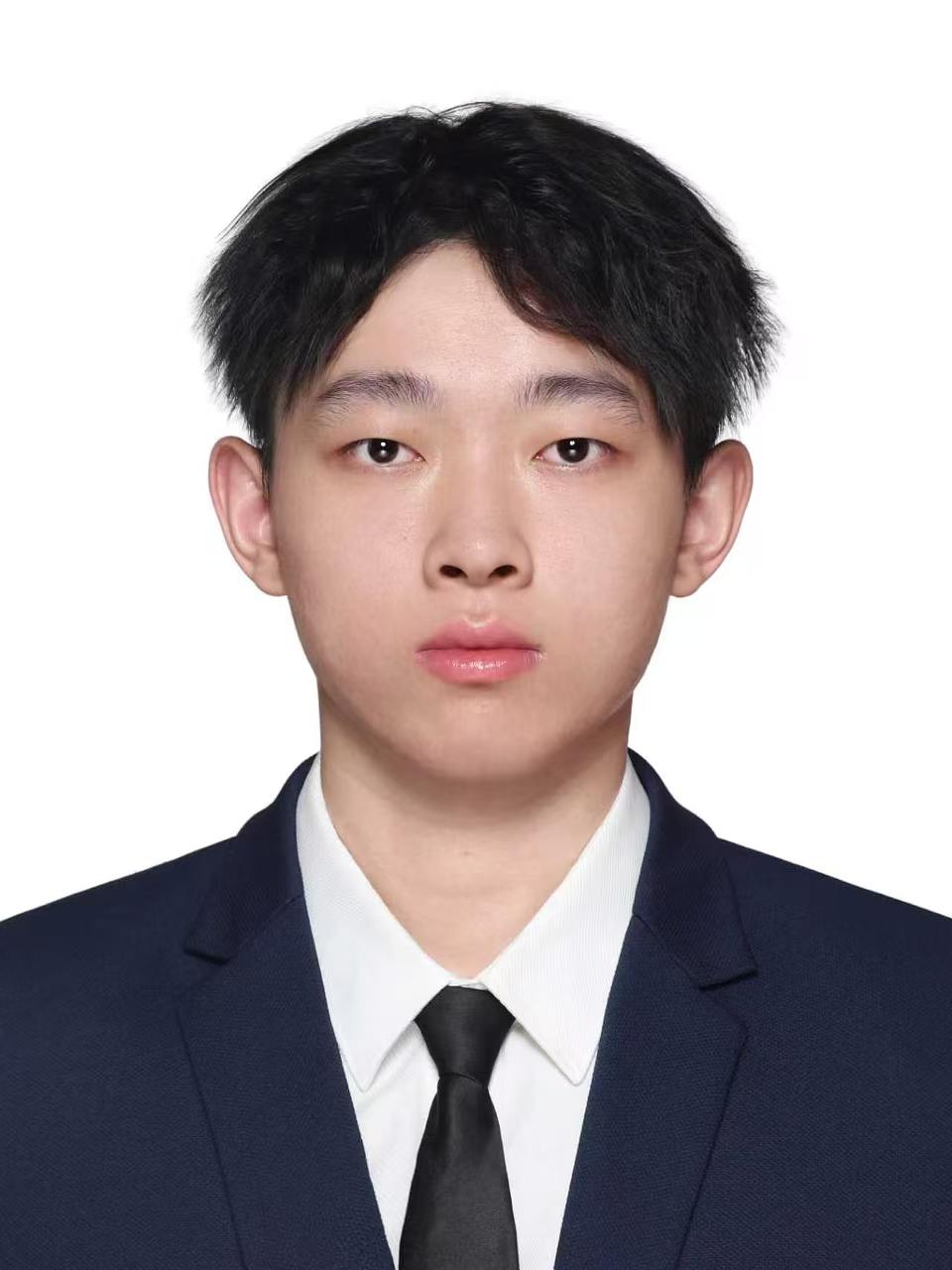}}]
{Haotian Guo} received the B.Eng. degree in microelectromechanical systems engineering from Northwestern Polytechnical University, Xi'an, China, in 2026. He is currently pursuing a master's degree in mechanical engineering at the National University of Singapore, Singapore.  His research interests include tactile sensing, sensor design, and robotic perception.
\end{IEEEbiography}

\begin{IEEEbiography}
[{\includegraphics[width=1in,height=1.25in]{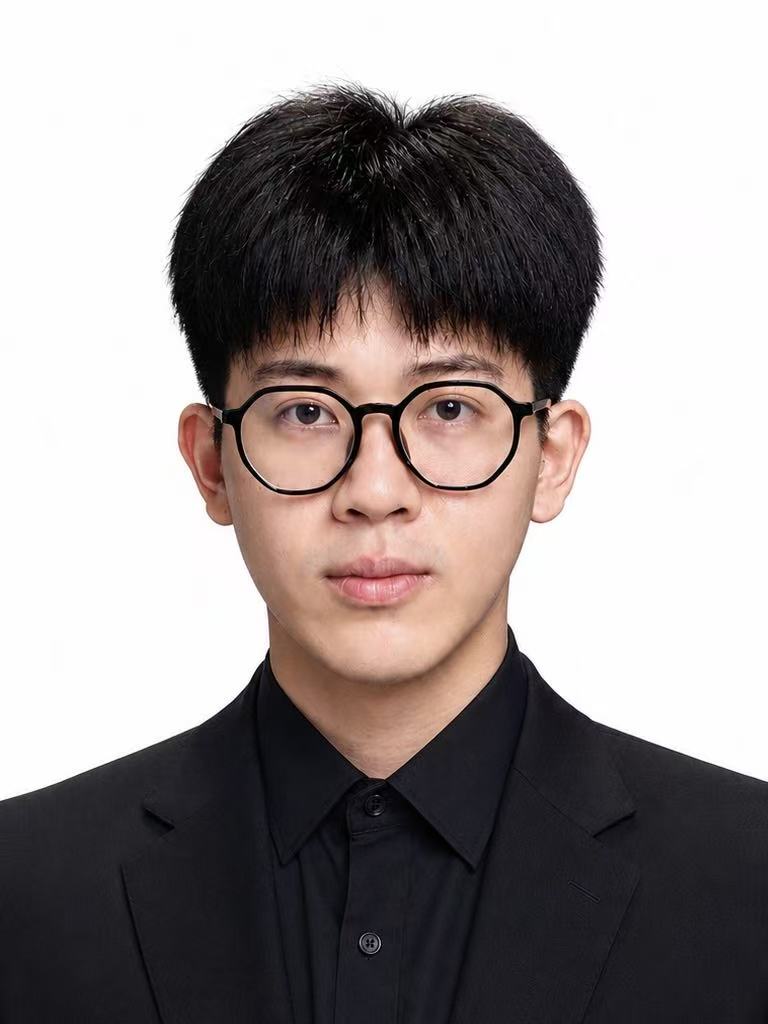}}]
{Haoming Liu} received the B.Sc. degree in optoelectronic information science and engineering from Sun Yat-sen University in 2026. He is currently working with the G\&M Lab, National University of Singapore, focusing on robot control, robotic manipulation, and multimodal perception.
\end{IEEEbiography}

\begin{IEEEbiography}
[{\includegraphics[width=1in,height=1.25in]{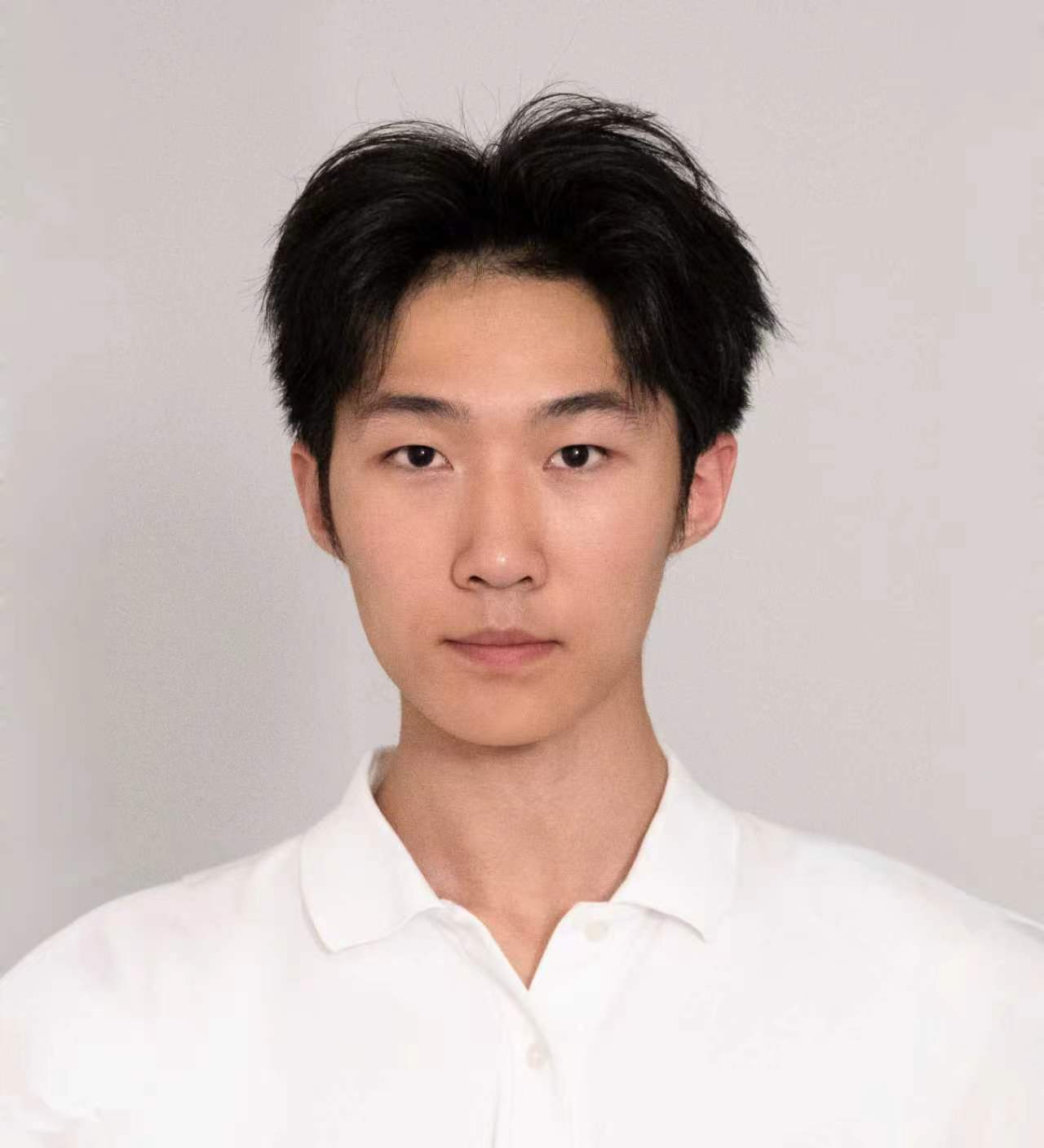}}]
{William Su} (Student Member, IEEE) is currently working toward the B.S. degree in aerospace engineering at the University of California, Berkeley, Berkeley, CA, USA. His research interests include dexterous manipulation, robotic end-effector design, and multimodal sensing.
\end{IEEEbiography}

\begin{IEEEbiography}
[{\includegraphics[width=1in,height=1.25in]{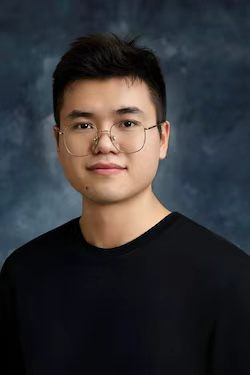}}]
{Jingxiang Guo} is a Ph.D. student at the National University of Singapore, in the Language and Vision Lab (LV-Lab), advised by Prof. Shuicheng Yan and co-supervised by Prof. Jianshu Zhou (G\&M Lab). He received his M.Comp. in AI from NUS and his B.Eng. in Automation from HITSZ.
\end{IEEEbiography}

\begin{IEEEbiography}[{\includegraphics[width=1in,height=1.25in,clip,keepaspectratio]{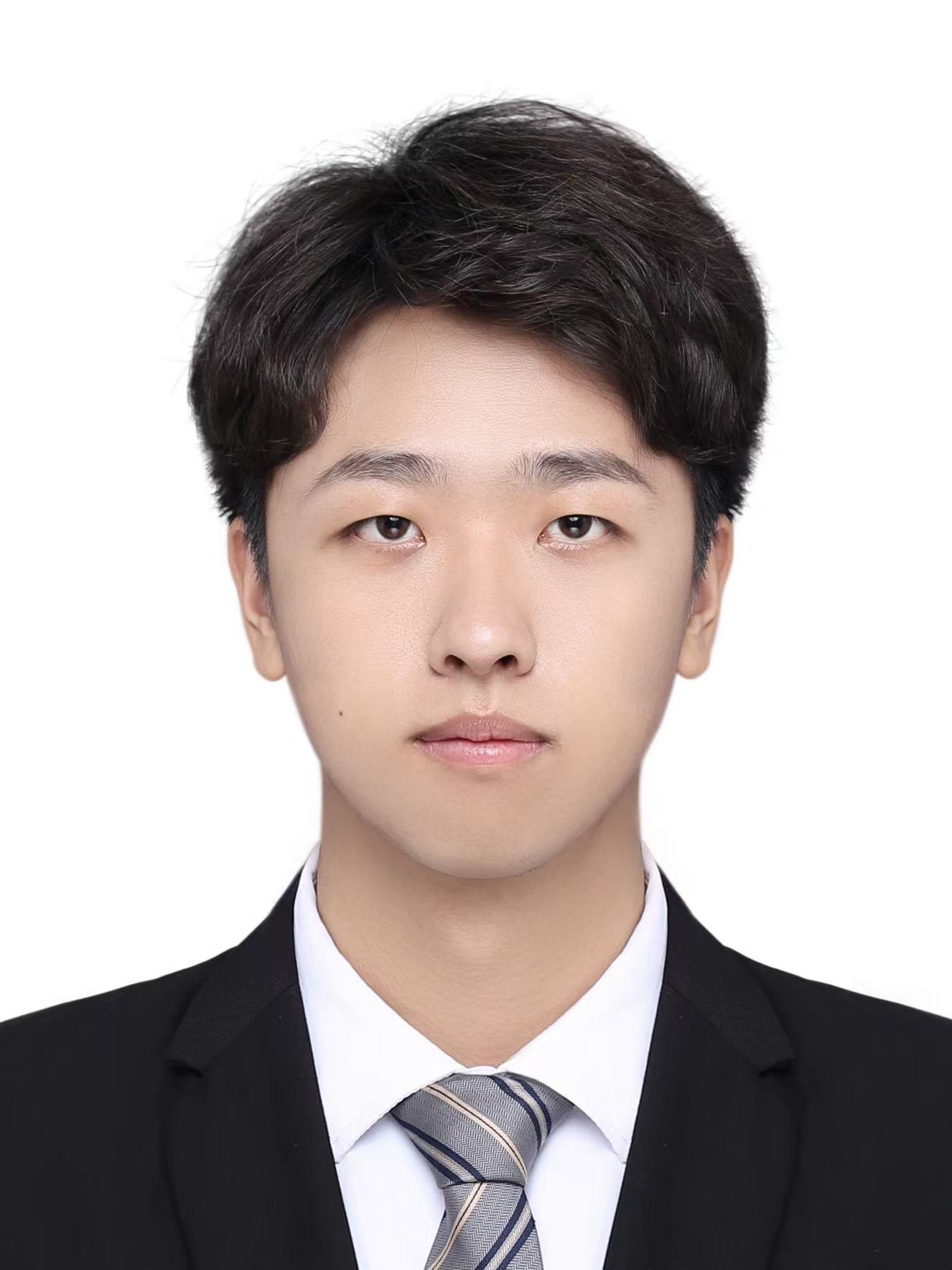}}]{Jiankun Li}
(Student Member, IEEE) received the bachelor's degree in automation engineering from Nanjing University of Aeronautics and Astronautics, Nanjing, China, in 2023, and the master's degree in electronic engineering from the University of Hong Kong, Hong Kong SAR, in 2024. He is currently pursuing a Ph.D. degree with the Department of Mechanical Engineering, City University of Hong Kong, Hong Kong SAR. His current research interests include tactile sensors and artificial intelligence.
\end{IEEEbiography}

\begin{IEEEbiography}[{\includegraphics[width=1in,height=1.25in,clip,keepaspectratio]{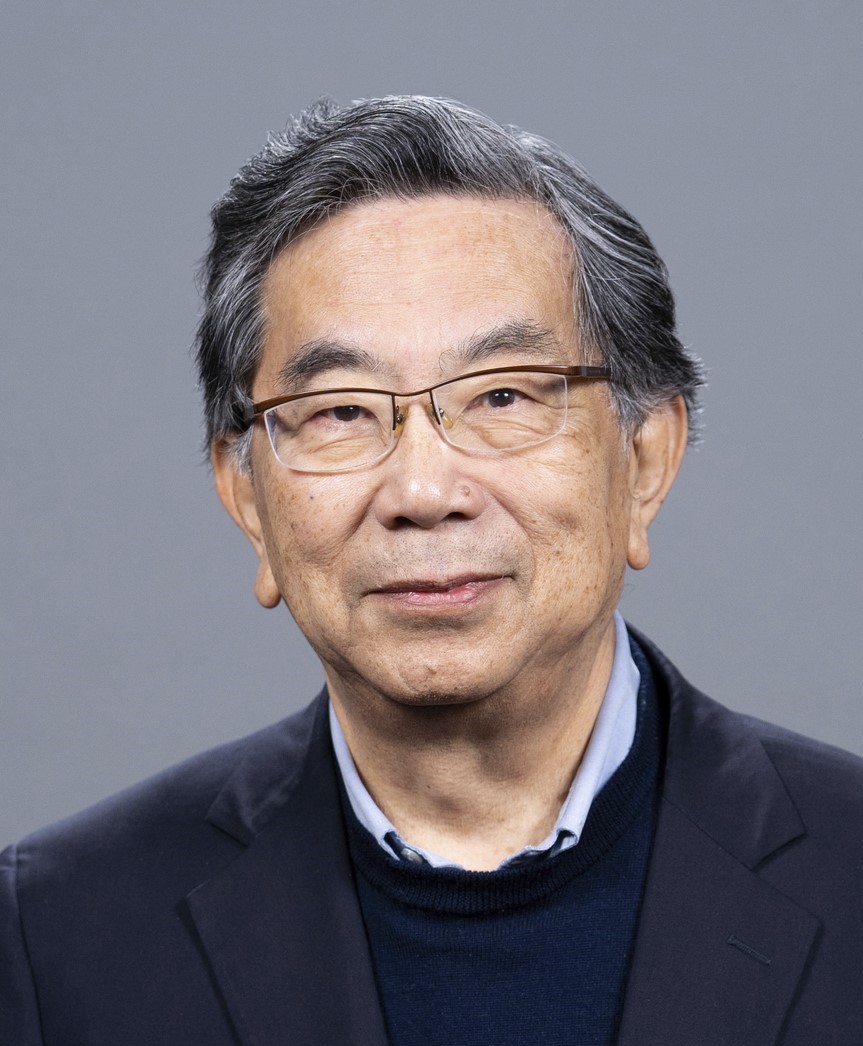}}]{Masayoshi Tomizuka} (Life Fellow, IEEE) was born in Tokyo, Japan, in 1946. He received the B.S. and M.S. degrees in mechanical engineering from Keio University, Tokyo, Japan, in 1968 and 1970, respectively, and the Ph.D. degree in mechanical engineering from the Massachusetts Institute of Technology, Cambridge, MA, USA, in February 1974. In 1974, he joined the Faculty of the Department of Mechanical Engineering, University of California at Berkeley, Berkeley, CA, USA, where he currently holds the Cheryl and John Neerhout, Jr., Distinguished Professorship Chair. At UC Berkeley, he teaches courses in dynamic systems and controls. From 2002 to 2004, he was the Program Director of the Dynamic Systems and Control Program of the Civil and Mechanical Systems Division of NSF. His current research interests are optimal and adaptive control, digital control, signal processing, motion control, and control problems related to robotics, machining, manufacturing, information storage devices, and vehicles. Dr. Tomizuka was the Technical Editor of the ASME Journal of Dynamic Systems, Measurement and Control, J-DSMC (1988--1993), Editor-in-Chief of the IEEE/ASME Transactions on Mechatronics (1997--1999), and Associate Editor for the Journal of the International Federation of Automatic Control, and Automatica. He was the General Chairman of the 1995 American Control Conference and was the President of the American Automatic Control Council (1998--1999). He is a Life Fellow of the ASME and a Fellow of the International Federation of Automatic Control (IFAC) and the Society of Manufacturing Engineers. He was the recipient of the J-DSMC Best Paper Award (1995, 2010), DSCD Outstanding Investigator Award (1996), Charles Russ Richards Memorial Award (ASME, 1997), Rufus Oldenburger Medal (ASME, 2002), John R. Ragazzini Award (AACC, 2006), Richard E. Bellman Control Heritage Award (AACC, 2018), Honda Medal (ASME, 2019), and Nathaniel B. Nichols Medal (IFAC, 2020). He is a member of the National Academy of Engineering.
\end{IEEEbiography}

\begin{IEEEbiography}[{\includegraphics[width=1in,height=1.25in,clip,keepaspectratio]{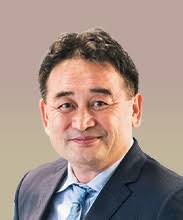}}]{Wen Jung Li}
(Fellow, IEEE) received the B.S. and M.S. degrees in aerospace engineering from the University of Southern California, Los Angeles, CA, USA, in 1987 and 1989, respectively, and the Ph.D. degree in aerospace engineering from the University of California at Los Angeles, in 1997. He worked in the Department of Mechanical and Automation Engineering, the Chinese University of Hong Kong (CUHK), from 1997 to 2011. Prior to CUHK, he held research and development positions with the NASA/Caltech Jet Propulsion Laboratory, Pasadena, CA, USA; The Aerospace Corporation, El Segundo, CA, USA; and Silicon Microstructures Inc., Fremont, CA, USA. He is honored as an International Fellow of The Canadian Academy of Engineering (CAE), a Member of the US National Academy of Artificial Intelligence (NAAI), a Fellow of European Academy of Engineering (EAE), an IEEE Fellow, an ASME Fellow, an Asia-Pacific AI Association (AAIA) Fellow, and a 100 Talents Program awardee of the Chinese Academy of Sciences. He served as the President of the IEEE Nanotechnology Council in 2016 and 2017. He currently holds a Chair Professorship of the Department of Mechanical Engineering and serves as the Vice-President (Talent and International Strategy) with the City University of Hong Kong, Hong Kong. His research interests encompass AI-enabled sensors, multimodal tactile sensing, and intelligent cyber-physical sensors.
\end{IEEEbiography}

\begin{IEEEbiography}
[{\includegraphics[width=1in,height=1.25in]{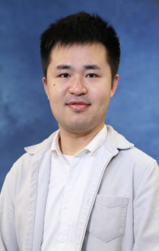}}]
{Jianshu Zhou} (Member, IEEE) received the Ph.D. degree in mechanical engineering from The University of Hong Kong, Hong Kong, in 2020. He is currently an Assistant Professor with the Department of Mechanical Engineering, National University of Singapore, Singapore. Prior to joining NUS, he held research positions at the University of California, Berkeley, and The Chinese University of Hong Kong. His research interests include robotics and mechatronics, robotic grasping and manipulation, embodied intelligence and physical AI, and multimodal sensing, data acquisition, and interaction. Dr. Zhou was the recipient of several international conference Best Paper Awards and Finalist recognitions and the ASME DSCD Rising Star Award in 2025. He currently serves as an Associate Editor for the IEEE Robotics and Automation Letters and a Technical Editor for the IEEE/ASME Transactions on Mechatronics. He is also a Member of ASME.

\end{IEEEbiography}
\end{document}